# A Review on Image Texture Analysis Methods

Shervan Fekri-Ershad[1,*]
[1]*Faculty of Computer Engineering, Najafabad Branch, Islamic Azad University, Najafabad, Iran*

[*]*Corresponding Author: fekriershad@pco.iaun.ac.ir*

## Abstract

Texture classification is an active topic in image processing which plays an important role in many applications such as image retrieval, inspection systems, face recognition, medical image processing, etc. There are many approaches extracting texture features in gray-level images such as local binary patterns, gray level co-occurence matrixes, statistical features, skeleton, scale invariant feature transform, etc. The texture analysis methods canbe categorized in 4 groups titles: statistical methods, structural methods, filter-based and modelbased approaches. In many related researches, authors have tried to extract color and texture features jointly. In this respect, combinated methods are considered as efficient image analysis descriptors. Mostly important challenages in image texture analysis are rotation sensitivity, gray scale variations, noise sensitivity, illumination and brightness conditions, etc. In this paper, we review most efficient and state-of-the-art image texture analysis methods. Also, some texture classification approaches are survived.

**Keywords:** Texture Classification, Local Binary Patterns, Impulse-noise, Tezxture Analysis, Color Texture Classification

# 1. Introduction

هدف اصلی این مقاله، مروری بر روش های انالیز بافت تصویر است. بنابراین در این بخش، ابتدا مفهوم بافت تصویر و تصویر بافتی بیان می گردد. سپس مسئله دسته‌بندی تصاویر بافتی تعریف می گردد. در ادامه صورت مسئله به تصاویر رنگی تعمیم داده شده و چالش های این حوزه مطرح خواهد شد. در نهایت اهمیت تحقیق در این حوزه مطرح خواهد شد.

## 1.1. Image Texture Definition

اصطلاح بافت تصویر، از تعریف بافت در پوشاک استنتاج شده است . در فرآورده های نساجی ، نوع، رنگ و چگونگی قرار گرفتن تار و پود در کنار یکدیگر بافت فرآورده را تشکیل می‌دهد. در پردازش تصویر و بینایی ماشین نیز به میزان، نوع و چگونگی پراکندگی و توزیع [1] شدت روشنایی [2] پیکسل‌ها در تمام طول تصویر در کنار یکدیگر بافت تصویر می گویند[1].

محققان در [2]، اصطلاح بافت تصویر را به صورت زیر تعریف کرده اند:

"یک حوزه بافتی در تصویر می‌تواند با یک توزیع فضایی نامنظم [3] ومتنوع از شدت روشنایی‌ها یا رنگ ها شناخته شود" .

ولیکن اصطلاح بافت تصویر و تصاویر بافتی با یکدیگر کاملاً متفاوت است. عبارت تصاویر بافتی [4] به تصاویری اطلاق می گردد که در آنها الگویی خاص از توزیع و پراکندگی شدت روشنایی پیکسل‌ها به صورت متوالی در تمام طول تصویر در حال تکرار شدن باشد.[1]

در شکل (1-1) چند نمونه از تصاویر بافتی نشان داده شده است . به طور مثال در این شکل الگوی خانه های سیاه و سفید در تصویر به صورت متوالی تکرار شده است و توالی آنها تصویر اصلی صفحه شطرنج را ساخته است.

---

[1] Distribution
[2] Intensity
[3] Non-uniform Spatial Distribution
[4] Texture Image

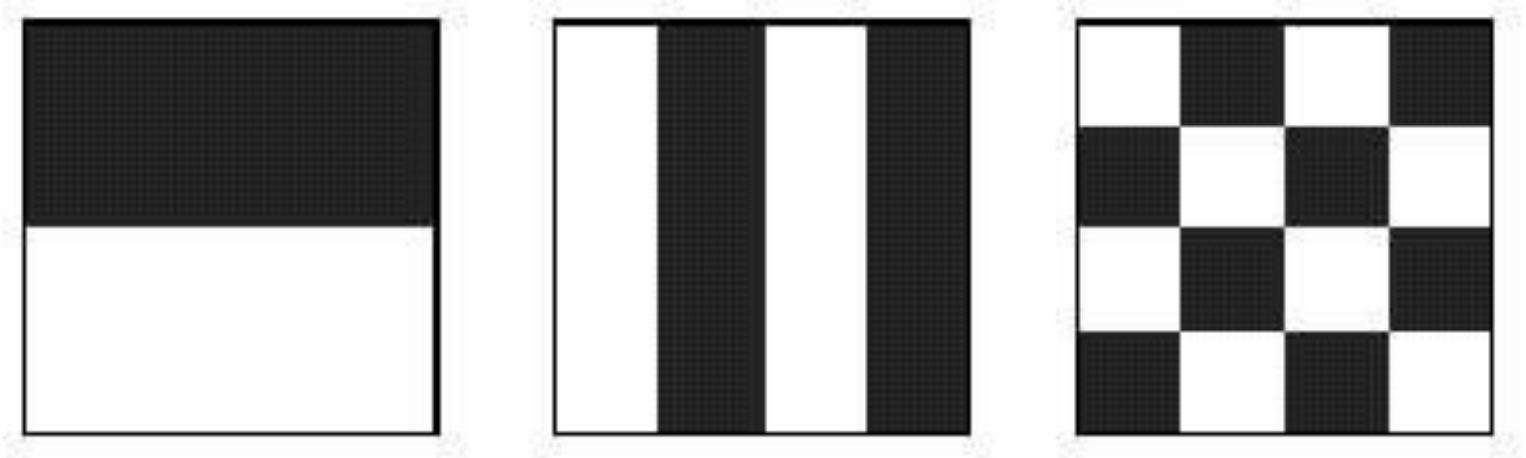

(چ) (ب) (الف)

شکل 1-1. نمونه هایی از تصاویر با بافت تکرارشونده [3]

در برخی از تصاویر ممکن است بخش های خاصی از تصویر دارای بافت تکرار شونده خاص خودشان باشند که با بخش های دیگر تفاوت می کند . به طور مثال تصویر نشان داده شده در شکل (1-2)، دارای سه چهار بافت مجزا شامل : بافت گلهای قرمز رنگ، بافت گلهای سفید رنگ، بافت چمن و بافت خیابان است[3]. چنین تصاویری را معمولا تصاویر بخشی بافتی [1] می نامند.

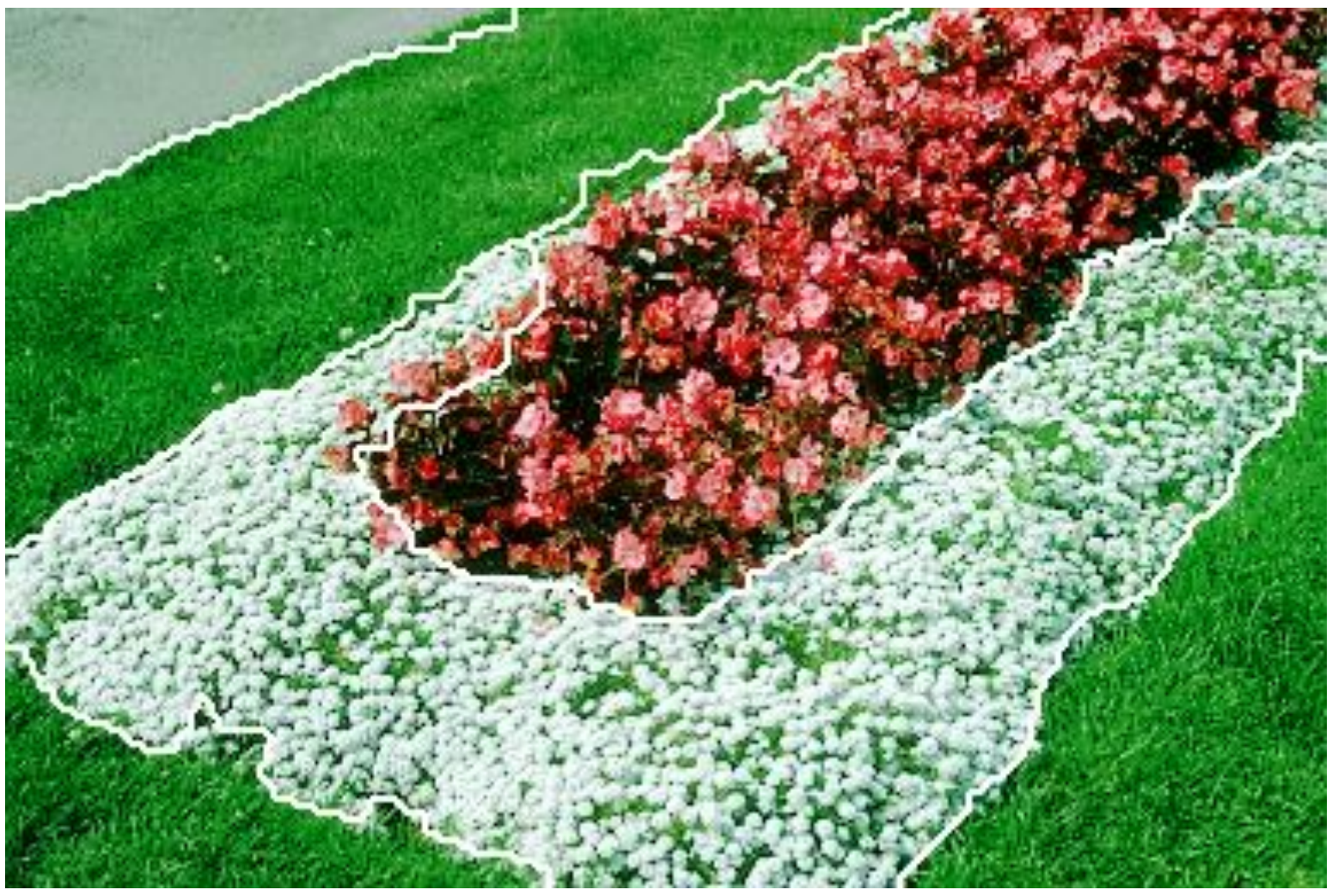

شکل 1-2. نمونه ای از تصویر بخشی بافتی با سه زیر بافت [3]

همانطور که قابل پیش بینی است، تصاویر بافتی یا توسط بشر تولید شدهاند و یا به صورت طبیعی در اطراف ما وجود دارند. نمونه هایی از این دو دسته تصویر بافتی در شکل زیر نشان داده شده است. به طور مثال در بخش الف شکل زیر، بافت طبیعی برگ درخت انگور، ساقه گندم، مزارع کشاورزی و کوهستان مشاهده می گردد. همچنین در شکل (1-3-ب) تصاویری از

[1] Regional Texture

ساختمان های مسکونی، فرآورده های حصیری، انواع پارچه و غیره مشاهده می شود، که همگی دارای بافت تکرار شونده بوده و در زمره تصاویر بافتی قرار می گیرند

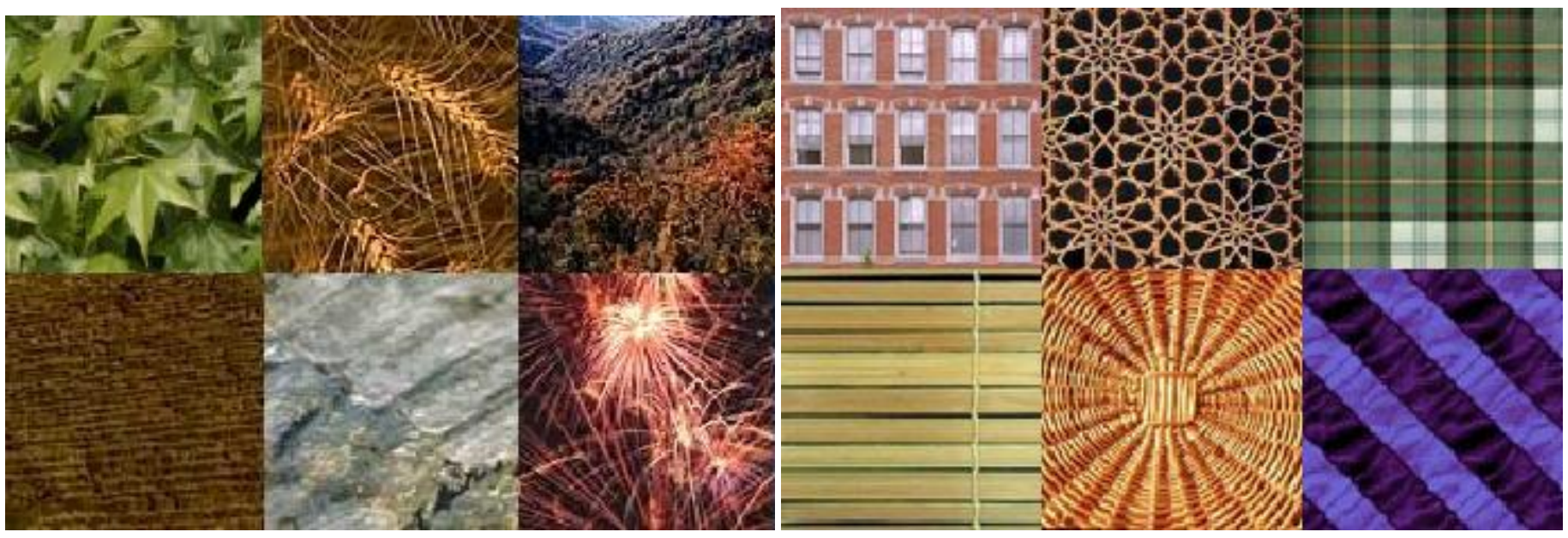

(الف) تصاویر با بافت هوشمندانه (ب) تصاویر با بافت طبیعی [4]

شکل 1-3. نمونه هایی از تصاویر طبیعی و هوشمندانه با بافت تکرار شونده

الگویی که در تصاویر بافتی در حال تکرار شدن است، را الگوی تکراری می نامند. به طور مثال در شکل (1-4-الف) تصویری از یک دیوار آجری گرفته شده است . همانطور که مشاهده می شود، بافت تصویر (همچنین بافت واقعی محیط ) کاملاً تکرار شونده می باشد به طوری که الگوی آجر به صورت متوالی در تمام سطح تصویر در حال تکرار شدن است . الگوی تکراری نیز در شکل (1-4-ب) به طور جداگانه بریده و نشان داده شده است. چند نمونه دیگر از تصاویری با بافت طبیعی[۱] تکرار شونده (چمنزار، تنه درخت) در شکل (1-5) نشان داده شده است. لازم به توضیح است که تصاویر مربوط به اشکال (1-4) و (1-5) از آلبوم بروداتز[۲] [5]، استخراج گردیده‌اند.

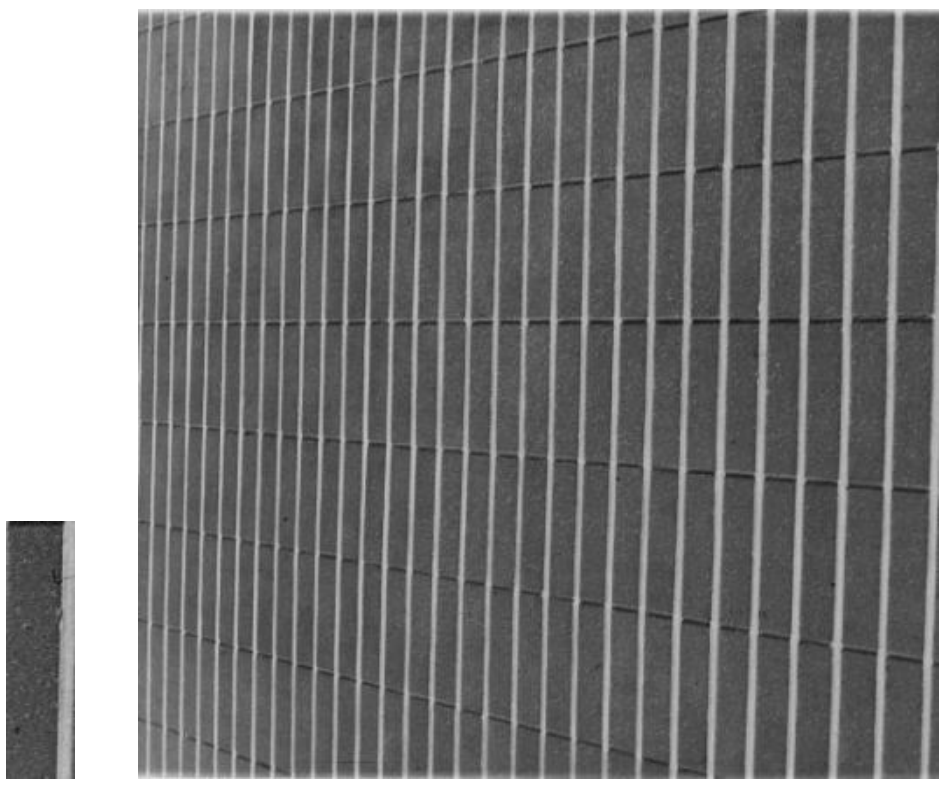

(الف) (ب)

شکل 1- 4. (الف)تصویری با بافت تکرار شونده از دیوار آجری (ب)الگوی تکراری در بافت تصویر(الف)[5]

[1] Natural Texture
[2] Brodatz

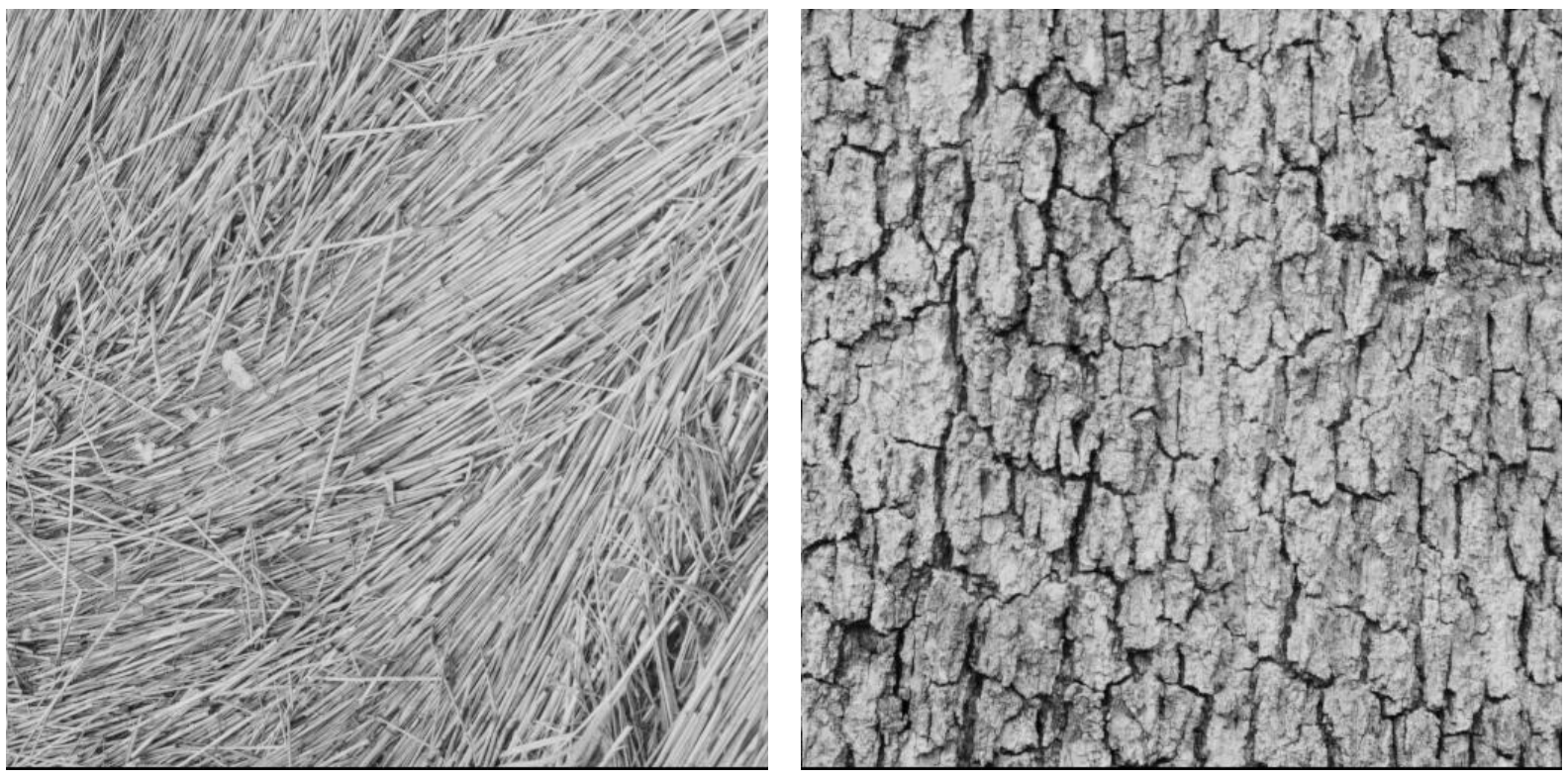

(ب) (الف)

شکل 1-5. چند نمونه از تصاویر طبیعی با بافت تکرار شونده [5]

(الف) تنه درخت (ب) چمنزار

## 1.2. Texture Analysis

بافت تصویر اطلاعات بسیار مفیدی در رابطه با اشیا[1] یا سطوح[2] به کار رفته در درون تصویر را با خود حمل می کند . از اواسط دهه 80، مفهومی به نام آنالیز بافت تصویر[3] مطرح شد. آنالیز بافت تصویر یکی از شاخه های مهم در پردازش تصویر است که برای تشخیص دادن یا تمییز دادن[4] سطوح و اشیا درون تصویر از یکدیگر استفاده می‌گردد. به صورت کلی، به روش هایی که یک سری ویژگی جهت تعریف بافت تصویر یا الگوی تکراری معرفی می کنند، عنوان روش‌های آنالیز بافت تصویر اطلاق می‌شود. در بینایی ماشین، آنالیز بافت می تواند به صورت مستقل و یا همراه با ویژگی‌های دیگر استخراجی از درون تصویر همچون رنگ، شکل[5]، حرکت[6] و ... استفاده شود[6]. بنابراین در اکثر روش های آنالیز تصویر، تلاش بر آن است تا یک بردار ویژگی که بتواند به بهترین نحو معرف بافت یا الگوی تکراری باشد، استخراج گردد. با توجه به این تفاسیر، از روش‌های آنالیز بافت تصویر در بسیاری از کاربردهای بینایی ماشین، همچون تشخیص الگو[7]

1 Objects
2 Surfaces
3 Texture Analysis
4 Discriminate
5 Shape
6 Motion
7 Pattern Recognition

بصری، پیگیری شی [۱]، تشخیص عیوب [۲]، سیستم های بازرسی بصری، شناسایی چهره [۳]، قطعه بندی تصویر[۴] و بازیابی تصویر[۵] استفاده می شود.

## 1.3. Texture Classification

در مفاهیم اولیه تشخیص الگو، عبارت دسته بندی[۶] به معنای نسبت دادن یک نمونه به یکی از گروه های از پیش تعریف شده است. در دسته بندی بافت تصوی[۷] هدف، نسبت دادن بافت یک تصویر نمونه ناشناس به یکی از کلاس های بافتی از پیش تعریف شده می باشد. در حوزه آنالیز بافت تصویر، معمولاً چهار مسئله اصلی مطرح می گردد که هر کدام حوزه های تحقیقاتی بسیار وسیعی می باشند. دسته بندی بافت تصویر یکی از این چهار زمینه تحقیقاتی در رابطه با آنالیز بافت تصویر می باشد. سه مسئله دیگر نیز عبارتند از بخش بندی بافت [۸]، ترکیب بافت[۹] و شکل بافت[۱۰].[8، 7]

دسته بندی بافت تصویر معمولاً شامل دو مرحله اصلی، استخراج ویژگی[۱۱] و مرحله تشخیص[۱۲] [7، 8، 9] است.

در مرحله اول، هدف ساختن یک مدل برای هر کدام از بافت‌هایی است که در تصاویر آموزشی پایگاه[۱۳] وجود دارند. ویژگی‌های استخراجی در این مرحله می توانند از نوع عددی، هیستوگرام-های گسسته، توزیع های تجربی [۱۴]، خصایص بافت [۱۵] همچون تباین، ساختار فضایی و جهت باشند. از آنجایی که هدف اصلی این مرحله آموزش بافت های مذکور به سیستم است، برخی محققان آنرا مرحله یادگیری[۱۶] نیز می نامند.

---

[1] Object Tracking
[2] Defect Detection
[3] Face Recognition
[4] Image Segmentation
[5] Image Retrieval
[6] Classification
[7] Image Texture Classification
[8] Texture Segmentation
[9] Texture Synthesis
[10] Shape of Texture
[11] Feature Extraction Phase
[12] Recognition Phase
[13] Training Data
[14] Empirical Distribution
[15] Texture Properties
[16] Learning Phase

در مرحله دوم، ابتدا بافت تصویر نمونه آزمایشی با همان روش به کار برده شده در مرحله قبل، آنالیز شده و سپس با استفاده از یک الگوریتم د سته بندی، ویژگی های استخراجی تصویر آزمایش با تصاویر آموزشی مقایسه شده و کلاس آن مشخص می گردد . در این مرحله با توجه به نوع ویژگی‌های استخراجی از انواع دسته بندی ها نیز می‌توان استفاده نمود. به همین دلیل این مرحله را دسته بندی[1] نیز می نامند.
تاکنون برای دسته‌بندی بافت تصویر، روش‌های متنوعی ارائه شده است. در اکثر این تحقیقات، محققان بر روی یکی از دو مرحله فوق متمرکز شده‌اند. شایان ذکر است که به دلیل گستردگی موضوع، تحقیقات در مورد مرحله آموزش نسبت به مرحله دسته بندی، بسیار بیشتر است.

## 1.4. Color Texture Classification

انسان معمولاً یک تصویر را به صورت ترکیبی از مولفه هایی همچون بافت، شکل و رنگ درک می‌کند[10]. بخش عمده ای از روش‌های آنالیز بافت تصویر ، فقط بر ای تصاویر سطوح خاکستری[2] تعریف شده اند. این در حالی است که یکی از ویژگی های کلیدی که می تواند در دسته بندی و آنالیز تصاویر استفاده شود، رنگ است . ترکیب ویژگی‌های رنگ تصویر در کنار ویژگی‌هایی همچون شکل و حرکت در کاربردهای گوناگونی از حوزه پردازش تصویر استفاده شده و نتایج مثبتی را به ارمغان آورده است که از آن جمله می توان به تشخیص اشیا [3][11] و ردیابی حرکت[4][12] اشاره کرد. بخش عمده‌ای از تصاویری که ما با آنها سر و کار داریم، رنگی هستند، بنابراین در نظر نگرفتن ویژگی های رنگ، در آنالیز چنین تصاویری به هیچ عنوان معقول نیست. اهمیت موضوع و تحقیقات نه چندان وسیعی که تاکنون در این حوزه صورت گرفته، ما را بر آن داشت تا در این مقاله، راهکاری برای دسته بندی بافت تصاویر رنگی ارائه نمایم. در راستای این هدف، در این مقاله تلاش خواهد شد که ویژگی‌های استخراجی از بافت و رنگ تصویر با یکدیگر ترکیب شوند و در نهایت ترکیب آنها میزان دقت دسته بندی را افزایش دهد.
در ترکیبِ ویژگی های بافت و رنگ، دو راهکار متداول وجود دارد : ترکیبِ اولیه [5] و ترکیبِ ثانویه[1] [13]. در راهکار ترکیب اولیه، عملگرهای آنالیز بافت بر روی هر کانال رنگی تصویر به

[1] Classification Phase
[2] Gray Levels Images
[3] Object Recognition
[4] Motion Tracking
[5] Early Fusion

صورت مجزا اعمال می شوند و در نهایت یک سری ویژگی های ترکیبی استخراج می گردد. این دسته از ویژگی‌های استخراجی زمانی که در تصویر ویژگی های بافت و رنگ در سطح پیکسل ترکیب شده باشند، قدرت جداسازی [۲] بالایی را فراهم کرده و دقت دسته بندی را افزایش می دهد. در نقطه مقابل، در راهکار ترکیب ثانویه، ویژگی های بافت و رنگ در سطح کل تصویر با یکدیگر ترکیب می شوند . در این دسته از روش ها، دو گروه ویژگی جداگانه (معمولا به شکل هیستوگرام) بر اساس رنگ و بافت استخراج شده و در نهایت این دو هیستوگرام مجزا به یکدیگر پیوست می‌شوند[۳] و بازنمایی[۴] نهایی را تشکیل می‌دهند [14]. در رابطه با جزییات هر کدام از روش های ترکیب فوق و متد های ارائه شده تاکنون با جزییات بیشتری در بخش آتی بحث خواهد شد.

## 1.5. Applications of Texture Analysis and Classification

با توجه به گستردگی وسیع تصاویر بافتی و بخشی بافتی، آنالیز بافت تصاویر رنگی یک مفهوم عمومی بوده که در بسیاری از مسائل حوزه پردازش تصویر و بینایی ماشین کاربرد وسیعی دارد. در اکثر کاربردها، محققین تلاش می کنند که از ویژگی های بافتی استخراجی از تصاویر (به صورت منفرد و یا در کنار ویژگی‌های دیگری همچون شکل، رنگ، حرکت و ...) برای جداسازی سطوح و اشیا استفاده نمایند . از جمله موارد کاربردی آنالیز و دسته بن دی بافت می توان به موارد زیر اشاره کرد:

الف – بازیابی تصویر[۵][15] : بازیابی تصویر به معنای استخراج تصاویر مشابه با تصویر پرس وجو از درون یک پایگاه بزرگ است . در این کاربرد، معیارهای بافتی و رنگ در کنار یکدیگر می توانند برای تعریف هر تصویر و اشیا درون آن استفاده گردند

ب – تشخیص عیوب سطحی[۶][16] : تشخیص عیوب به معنای مشخص کردن مکان و قسمت-هایی از سطح مورد نظر در تصویر است که در آن ن قاط عیب و نقص‌هایی رخ داده است. پر واضح است که در قسمت های معیوب سطوح، بافت تصویر نسبت به بافت بخش های سالم

[1] Late Fusion
[2] Discriminative
[3] Concatenating
[4] Representation
[5] Image Retrieval
[6] Surface Defect Detection

تفاوت کرده است . بنابراین آنالیز و دسته بندی بافت در این حوزه نیز کاربرد مهمی خواهد داشت.

ج – ردیابی اشیا[1] [17] : در این کاربرد هدف، تشخیص و ردیابی یک شی خاص در درون تمامی فریم های یک ویدیو است. بدون شک، می بایست در مرحله یادگیری، شی مورد نظر به سیستم معرفی گردد. در این مرحله ویژگی های بافتی و رنگ شی می توانند مولفه های بسیار جداپذیری باشند. از جمله موارد دیگر کاربردی دسته بندی بافت تصاویر رنگی می توان به تشخیص چهره[2] [18]، سیستم‌های بازرسی بصری[3][19] اشاره نمود.

# 2. Related Works in Texture Analysis

همانطور که در فصل قبل مشخص شد، یکی از ا هداف این مقاله. بررسی اجمالی روش های کارآمد در حوزه دسته بندی و آنالیز بافت تصویر است . بنابراین پس از بررسی صورت مسئله و اهمیت تحقیق در بخش پیشین، در این بخش به مرور تح قیقاتی که تاکنون در این حوزه صورت گرفته است، پرداخته می شود. مطالب این بخش را می‌توان به دو قسمت تقسیم نمود. ابتدا به گروه‌بندی روش‌های آنالیز بافت تصاویر و بررسی آنها، پرداخته میشود. سپس به مرور روش‌هایی می‌پردازیم که جهت دسته بندی تصاویر رنگی، بافت و رنگ تصویر را ترکیب نموده-اند.

## 2.1. A review on Image Texture Analysis methods

همانطور که در بخش قبل اشاره شد، هدف اصلی این مقاله ارائه روشی برای آنالیز بافت تصاویر رنگی است. بنابراین بهتر است که در این بخش، ابتدا به بررسی روش‌های آنالیز بافت تصویر پرداخته و سپس روش‌های ترکیبی بافت و رنگ مورد بررسی قرار گیرد. تاکنون محققین روش-های متفاوتی در حوزه آنالیز بافت ارائه کرده اند که هر کدام از آنها به تنهایی می تواند نکات مثبت و منفی متفاوتی داشته باشد . دسته بندی روش های ارائه شده با توجه به پیچیدگی

[1] Object Tracking
[2] Face Recognition
[3] Visual Inspection Systems

موضوعات چندان کار آسانی نیست، ولی به طور ضمنی می توان روش های ارائه شده در این زمینه را به 4 گروه کلی به شرح زیر تقسیم کرد[20 و21].

- روش های آماری[۱]
- روش های ساختاری[۲]
- روش های مبتنی بر فیلتر[۳]
- روش های مبتنی بر مدل[۴]

در ادامه این بخش، ابتدا به تعریف مفهوم هر کدام از گروه های بالا پرداخته شده و سپس برای هر کدام چندین روش به اختصار بررسی می شود. سپس برخی از روش‌های مطرح جهت آنالیز بافت که به طور مستقیم به این مقاله مربوط می شوند، به تفصیل بررسی می‌گردد.

### 2-1-1- روش های آنالیز آماری

این دسته از روش‌ها جهت آنالیز بافت تصاویر، یک سری محاسبات آماری بر روی توابع توزیع شدت روشنایی پیکسل ها انجام می دهند . به طور کلی روش‌هایی که برای استخراج بردار ویژگی معرف بافت تصویر، از محاسبات آماری و ریاضی استفاده می کنند، در زمره‌ی این گروه قرار می‌گیرند. از جمله روش‌های این گروه می توان به خصایص هیستوگرام[۵][22، 23 و 24]، ماتریس‌های هم رخدادی[۶][25، 26 و 27]، الگوی دودویی محلی[۷][17 و 28] و تابع همبستگی[۸][29 و 30] اشاره کرد. برای درک بهتر مفهوم روش‌های آماری، روش ماتریس های همرخدادی و خصایص هیستوگرام به تفصیل شرح داده خواهد شد.

**الف- ویژگی ها و خصایص هیستوگرام**

هیستوگرام تصویر، نمایشی دو بعدی از چگونگی پراکندگی درجات سطوح خاکستری در تصویر است. به طوری که بعد افقی آن نشان دهنده سطوح خاکستری به کار رفته در تصویر و بعد عمودی هم نشان دهنده تعداد پیکسل هایی از تصویر با سطح خاکستری مورد نظر است . به طور مثال، هیستوگرام تصویری 3 سطحی با 16 پیکسل در شکل (3-1) نشان داده شده است.

---

1 Statistical
2 Structural
3 Filter Based
4 Model Based
5 Histogram Properties
6 Co-Occurrence Matrix
7 Local Binary Pattern
8 Auto-Correlation

| | | | |
|---|---|---|---|
| **2** | **2** | **0** | **2** |
| **2** | **2** | **0** | **1** |
| **0** | **1** | **2** | **0** |
| **2** | **2** | **0** | **1** |

(الف)

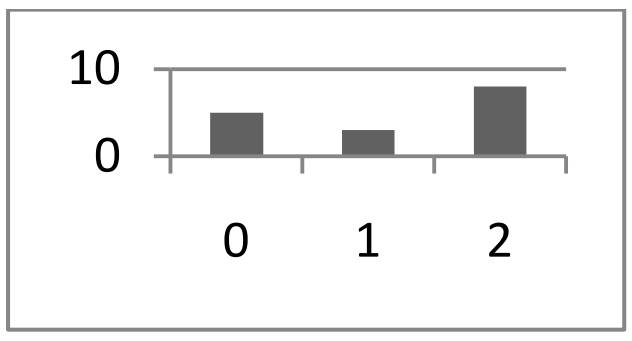

(ب)

شکل2-1. (الف) تصویری با 3 سطح خاکستری (ب) هیستوگرام تصویر (الف)

در شکل (2-2) نیز هیستوگرام دو تصویر واقعی نشان داده شده است. لازم به توضیح است که هر دو تصویر دارای 256 سطح خاکستری هستند.

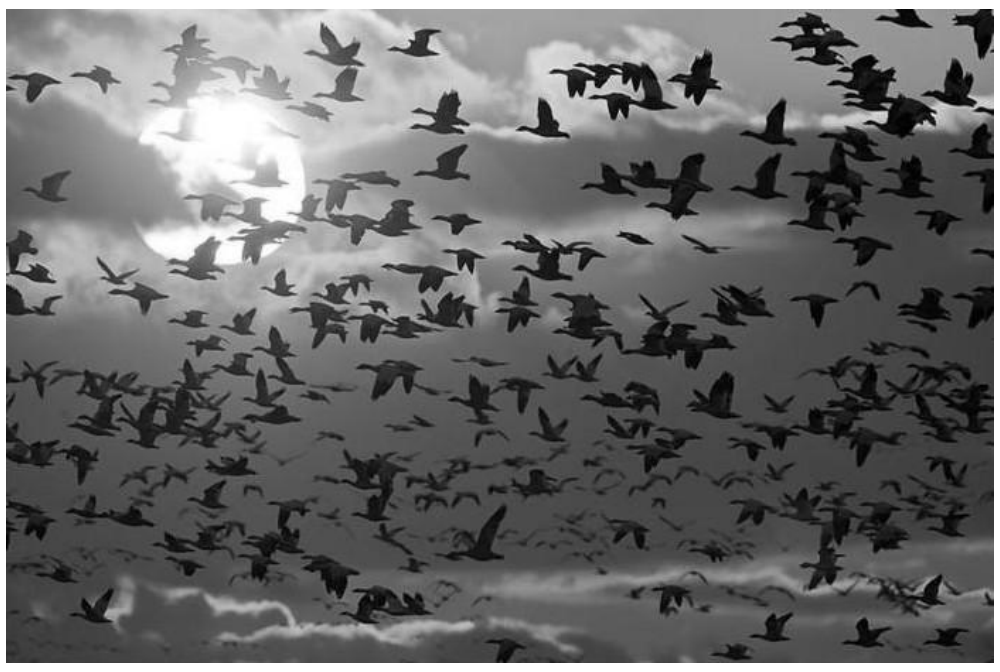

(الف)

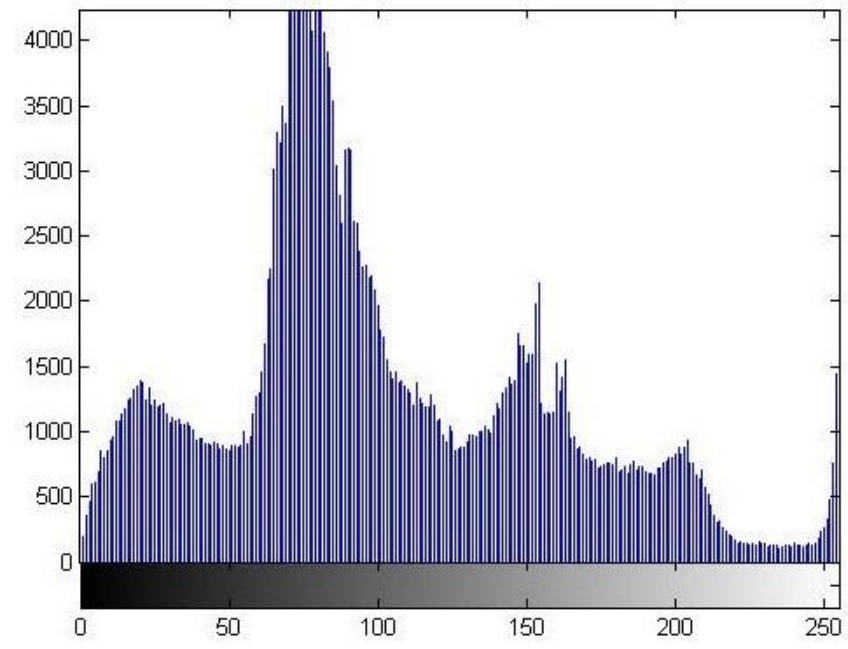

( ب )

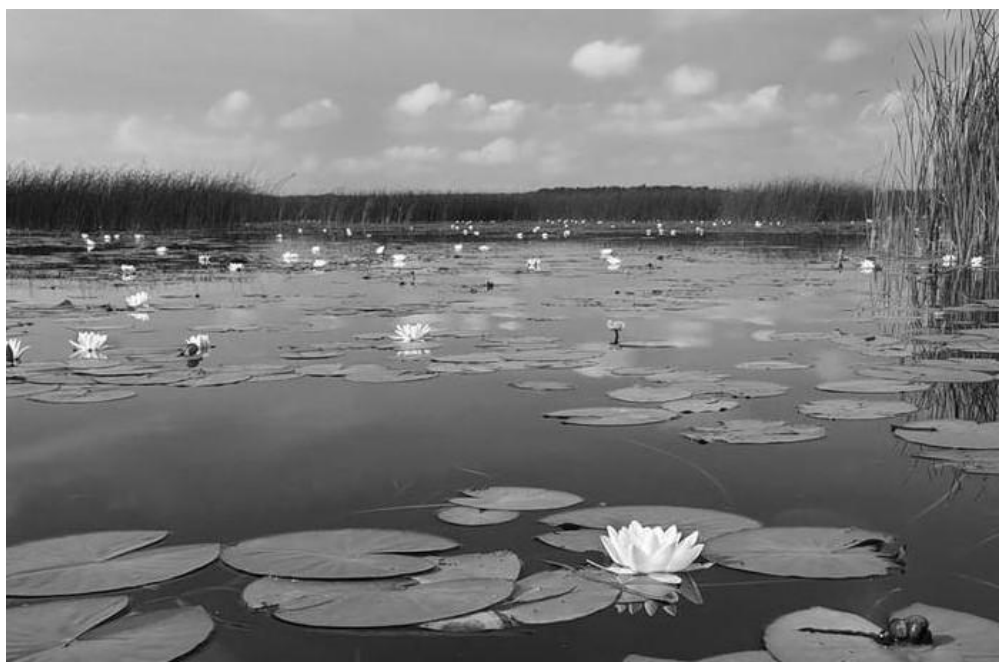

(ج)

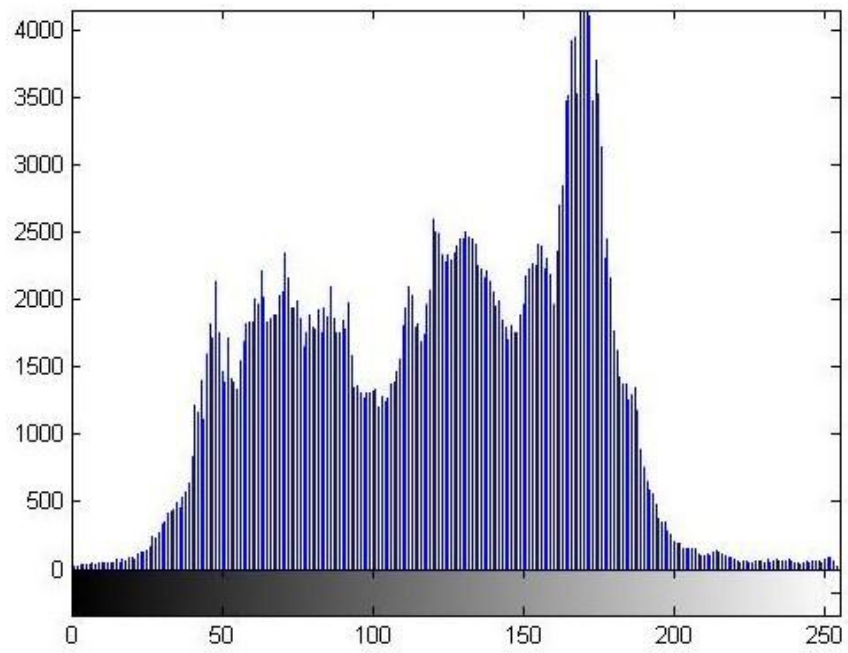

( د )

شکل 2-2. نمونه هایی از هیستوگرام

(الف) تصویر اصلی [4]، (ب) هیستوگرام تصویر الف، (ج) تصویر اصلی[4]، (د) هیستوگرام تصویر ج

هیستوگرام تصویر می‌تواند به صورت مستقیم یا غیر مستقیم، اطلاعات با ارزشی از تصویر را در اختیار کاربر قرار دهد. در همین راستا یک سری معیارهای آماری که به ویژگی های "هارلیک"

$$Mean = \bar{g} = \sum_{g=0}^{L-1} gP(g) \tag{10}$$

نیز شهرت دارند، ارائه شده‌اند که هر کدام خصوصیاتی از تصویر را نشان می دهند. برخی از این ویژگی ها به شرح زیر هستند(معادلات 2-1 الی 2-4):

(1 - 2)

معادله (2 - 1)، میانگین تصویر را محاسبه می‌کند که می تواند معرف خوبی از میزان تابناکی[1] تصویر باشد. در این معادله g نشان دهنده سطوح خاکستری بوده و P(g) احتمال رخداد سطح g در تصویر است.

$$\sigma_g = \sqrt{\sum_{g=0}^{L-1} (g-\bar{g})^2 P(g)} \quad (2 - 2)$$

معادله (2 - 2)، انحراف از استاندارد تصویر را محاسبه می کند که می تواند معرف خوبی از میزان وسعت اطلاعات[2] در تصویر باشد.

$$E = \sum_{g=0}^{L-1} [P(g)]^2 \quad (3 - 2)$$

معادله (2 - 3)، انرژی را در تصویر محاسبه می‌کند که می‌تواند معرف خوبی از چگونگی توزیع سطوح خاکستری[3] در تصویر باشد.

$$H = -\sum_{g=0}^{L-1} P(g) \log_2 [P(g)] \quad (4 - 2)$$

معادله (2 - 4)، آنتروپی تصویر را محاسبه می‌کند که می تواند معرف خوبی از میزان بیت‌های لازم جهت کد کردن تصویر باشد.

در معادلات (2 - 1) الی (2 - 4)، L نشان دهنده ماکزیمم سطوح خاکستری و P(g) نشان دهنده احتمال برخورد به هر کدام از سطوح خاکستری (ارتفاع آن سطح در هیستوگرام نرمال شده) است. در نهایت میزان هر کدام از این معیارها در تصویر مورد نظر می تواند معرف خوبی از بافت آن تصویر باشد . بنابراین برای آنالیز بافت تصاویر می توان از خصایص هیستوگرام استفاده نمود.

### ب- ماتریس‌های هم‌رخدادی

یکی از عملگرهای پرکاربرد برای آنالیز بافت تصاویر، ماتریس های هم رخدادی هستند. در اصطلاح ماتریسهای هم رخدادی، چگونگی وابستگی فضایی سطوح خاکستری مختلف نسبت به

---
[1] Brightness
[2] Spread
[3] Gray-Level Distribution

یکدیگر را می سنجند . در همین راستا ابتدا نوع خاصی از رابطه فضایی تعریف شده و سپس میزان رخداد رابطه فضایی مورد نظر برای هر کدام از سطوح خاکستری درون تصویر نسبت به تمامی سطوح دیگر بررسی می شود. برای فهم بهتر، مثالی در شکل (2-3) نشان داده شده است. در این شکل، ابتدا نوع رابطه فضایی مورد نظر توسط Relation مشخص شده و سپس میزان رخداد سطوح خاکستری نسبت به یکدیگر (با در نظر گرفتن رابطه مورد نظر) در ماتریس محاسبه شده است . توجه گردد که رابطه نشان داده شده به معنای رابطه نود درجه (یک پیکسل به سمت پایین ) است. همانطور که می بینید هر کدام از سطور افقی مربوط به یک سطح بوده و هر خانه از ماتریس نشان می دهد که چندبار این سطح با سطح خاکستری ستون مورد نظر دارای رابطه فضایی R بوده است. به طور مثال خانه سطر سوم و ستون اول ماتریس نشان می دهد که 4 بار پس از سطح خاکستری 2 در تصویر در پیکسل سمت راست سطح خاکستری صفر وجود دارد.

| | | | |
|---|---|---|---|
| 2 | 2 | 0 | 2 |
| 2 | 2 | 0 | 1 |
| 0 | 1 | 2 | 0 |
| 2 | 2 | 0 | 1 |

(الف)

| | 0 | 1 | 2 |
|---|---|---|---|
| 0 | 1 | 1 | 2 |
| 1 | 1 | 0 | 1 |
| 2 | 2 | 2 | 2 |

R =

| |
|---|
| 0 |
| 1 |

(ب)

شکل 2-3. (الف) تصویر اصلی با 3 سطح خاکستری (ب) ماتریس هم رخدادی برای (الف) با رابطه R

با توجه به توضیحات فوق، می توان با توجه به هر نوعی از رابطه فضایی، یک ماتریس هم رخدادی برای تصویر مورد نظر محاسبه کرد . از آنجایی که ماتریس های هم رخدادی ارتباطات پیکسل‌ها با یکدیگر را نشان می دهند، بنابراین چنانچه بتوان معیارهای با ارزشی از درون آنها استخراج نمود، این ماتریس ها می‌توانند معرف خوبی از بافت تصویر مورد نظر باشند . در این راستا تاکنون معیارهای آماری گوناگونی تعریف شده که از اصلی ترین آنها می توان به انرژی و آنتروپی و تباین[1]، هم جنسی [2] و همبستگی[3] اشاره کرد. برای محاسبه این سری از ویژگی ها

---
[1] Contrast
[2] Homogeneity
[3] Correlation

باید در ابتدا ماتریس هم رخدادی نرمال شده و سپس معادلات مربوطه محاسبه شود . برای نرمال سازی ماتریس کافی است که ارزش هر خانه به مجموع کل ارزش های درون ماتریس تقسیم گردد. معادله (2 -5 ) این موضوع را بیان می کند.

$$N(p,q) = V(p,q) / \sum_{i=1}^{n} \sum_{j=1}^{m} V(i,j) \qquad (5\text{-}2)$$

در معادله فوق، mوn نشان دهنده سایز ماتریس همرخدادی هستند که برابر با تعداد سطوح خاکستری به کار رفته در تصویر است. همچنین p وq به ترتیب سطر و ستون خانه مورد نظر، V ماتریس هم رخدادی و N ماتریس هم رخدادی نرمال شده را نشان می دهند. نحوه محاسبه معیارهای آماری فوق در معادلات (3-6) الی (3-10) با جزییات بیان شده است[6 , 3].

$$H = -\sum_{i=1}^{n} \sum_{j=1}^{m} N(i,j) Log(N(i,j)) \qquad (6\text{-}2)$$

$$Constant = \sum_{i=1}^{n} \sum_{j=1}^{m} (i-j)^2 N(i,j) \qquad (7\text{-}2)$$

$$E = \sum_{i=1}^{n} \sum_{j=1}^{m} N^2(i,j) \qquad (8\text{-}2)$$

$$Homogeneity = \sum_{i=1}^{n} \sum_{j=1}^{m} {N(i,j)} \Big/ {1+|i-j|} \qquad (9\text{-}2)$$

$$Correlation = \frac{\sum_{i=1}^{n} \sum_{j=1}^{m} (i-\mu_i)(j-\mu_j) N(i,j)}{\sigma_i \sigma_j} \qquad (10\text{-}2)$$

## 2-1-2- روش های آنالیزساختاری

جهت آنالیز بافت تصویر، برخی از روش‌ها ساختار بافت را مورد بررسی قرار می دهند . بدین معنا که الگوها و بافت‌هایی را از پیش طراحی کرده و به دنبال میزان و مناطق رخداد آنها در تصویر می‌گردند. ساختارهای از پیش طراحی شده می توانند بسیار ساده باشند همچون ویژگی خاصی برای هر پیکسل و یا تا حدی پیچیده تر باشند همچون ساختارهایی برای یک همسایگی کوچک یا بزرگ. البته در برخی دیگر از روش های این گروه نیز ویژگی های بافتی (معیارهایی عددی) برای شناسایی استفاده می‌شوند. ازجمله مشهورترین روش هایی که در زمره این گروه جای می گیرند می توان به سنجش نمایش اسکلت[1] [31 و 32]، عملگر های مورفولوژی[2] [33 و

---

[1] Skeleton Representation
[2] Morphological Operators

34]، واحد های ساختاری پیش تعریف[۱] [35 و 36]، ویژگی‌های لبه[۲] [37، 38 و 39]، اشاره کرد. در ادامه برای درک بهتر مفهوم ساختار و روش های ساختاری جهت نمونه روش ویژگی - های لبه با جزییات مورد بررسی قرار خواهد گرفت.

### الف- ویژگی های لبه

به پیکسل‌هایی از تصویر که شدت روشنایی آنها نسبت به همسایگانش جهش قابل ملاحظه‌ای داشته باشد(افزایش/ کاهش) اصطلاحاً لبه گفته می شود. لبه‌یابی یکی از مسائل پرکاربرد در حوزه پردازش تصویر است . این عمل توسط اپراتورهای مختلفی همچون سوبل[۳]، کنی[۴]، پریویت[۵] و ... می‌تواند صورت گیرد. عملکرد این اپراتورها بدین صورت است که به وابسته تعریف نوعی فیلتر، در تصویر به دنبال پیکسل هایی می گردند که می تواند شروط فیلتر را ارضا کند . (فیلترها به صورتی طراحی شده اند که تنها پیکسل هایی که جزو لبه هستند، شرایط را ارضا می کنند). چند نمونه از تصاویر لبه یابی شده توسط فیلترهای مختلف در شکل (3 - 4) نشان داده شده است.

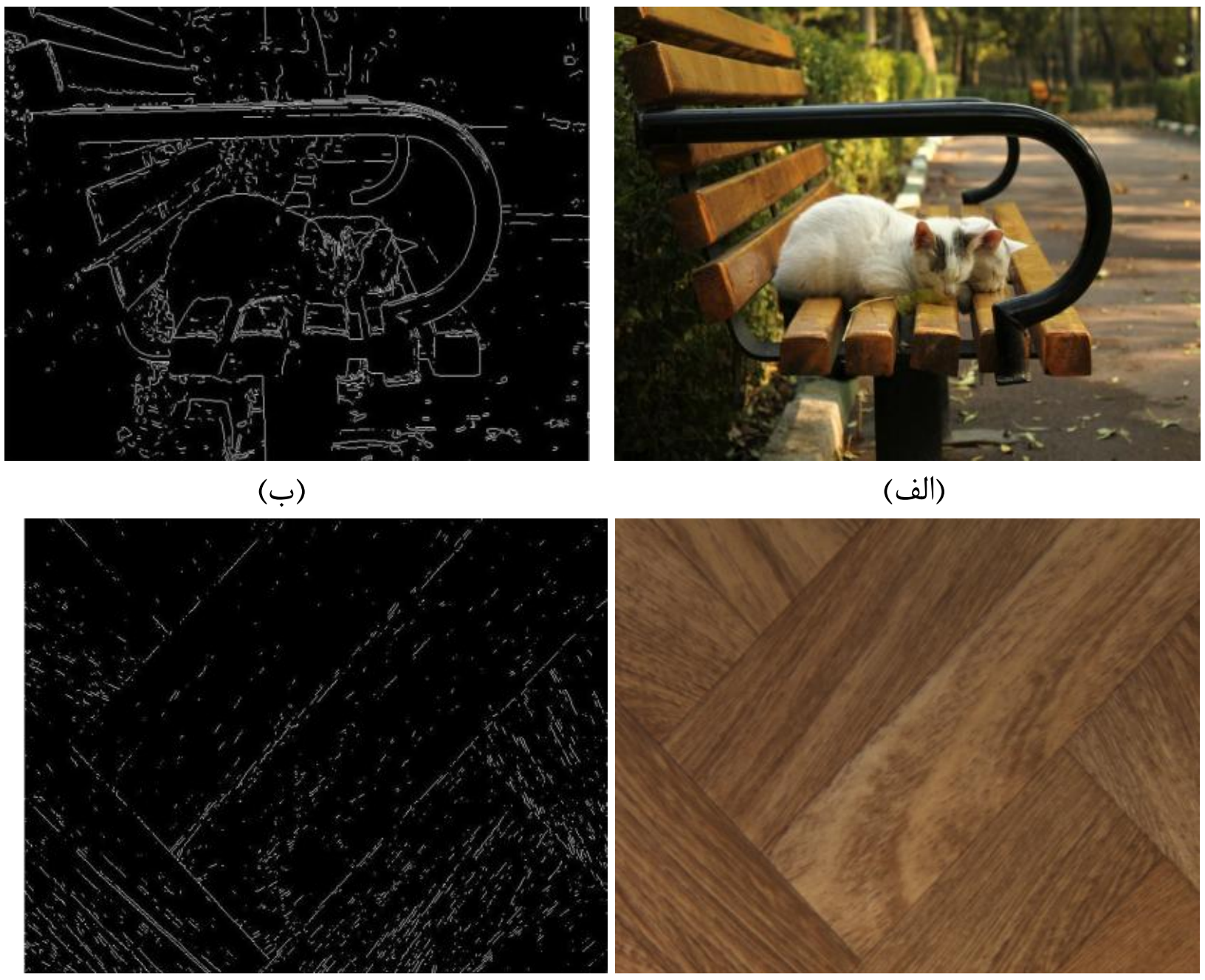

(الف) (ب)

(ج) (د)

شکل 2-4. نمونه هایی از لبه یابی

(الف) تصویر اصلی [109]، (ب) تصویر لبه یابی شده توسط فیلتر سوبل با حد آستانه 0.07

(ج) تصویر اصلی [109]، (د) تصویر لبه یابی شده توسط فیلتر کنی با حد آستانه 0.3

---

[1] Primitive Measurement
[2] Edge Features
[3] Sobel
[4] Canny
[5] Prewitt

تصویر لبه یابی شده می تواند معرف خوبی برای بافت تصویر باشد، به شرط آنکه بتوان ویژگی های قابل محاسبه‌ای را از درون آن استخراج نمود. در همین راستا ویژگی هایی همچون تعداد پیکسل های لبه، هیستوگرام بزرگی [1] و هیستوگرام جهت [2]، تاکنون ارائه شده‌اند. همانطور که از تعریف فیلتر لبه یابی و عملکرد آن برداشت می‌شود، این فیلترها در تصویر به دنبال ساختارهای معینی می‌گردند، بنابراین می توان این روش را در زمره روش های ساختاری قرار داد.

## ب- Scale Invariant Feature Transform (SIFT)

دیوید لوو در [136]، روشی خلاقانه برای استخراج ویژگی هایی محلی با قدرت تفکیک دهندگی [3] بالا از تصویر، تحت عنوان "تبدیل ویژگی های مستقل از مقیاس (SIFT)" ارائه داد. روش ارائه شده در [136] که برای حل مسئله تشخیص شی [4] مطرح شده است، تصویر را به مجموعه بزرگی از بردارهای ویژگی منتقل می کند که هر کدام از آنها نسبت به چرخش و جابجایی تصویر [5] کاملا مستقل بوده و نسبت به تغییرات میزان روشنایی [6] تصویر نیز مقاومت بالایی دارند. نحوه استخراج ویژگی توسط عملگر SIFT را می توان یک روش مرحله‌ای دانست که در مرحله اول ابتدا مکان های کلیدی [7] تصویر شناسایی می شوند. در [136]، مکان های کلیدی تصویر، نقاطی از تصویر تعریف می شوند که تابع تفاوت گوسین [8] (DoG) در آن نقاط ماکزیمم یا مینیمم می‌شوند. سپس هر نقطه کلیدی برای تولید یک بردار ویژگی استفاده می شود که بخش‌های محلی [9] تصویر را توصیف می نمایند. دیود لوو در [136]، بردارهای ویژگی استخراجی را نقاط SIFT نامیده و اعتقاد دارد که این ویژگی ها نسبت به تغییرات محلی تصویر تا حد زیادی مستقل هستند. دیوید لوو در [137]، روش مبتکرانه خود را با ترکیب تشخیص دهنده ناحیه مستقل از مقیاس و عملگری مبتنی بر نحوه توزیع گرادیان محلی [10] تصویر بهبود می بخشد.

---

[1] Magnitude
[2] Direction
[3] Discriminative
[4] Object Recognition
[5] Image Translation
[6] Illumination
[7] Key locations
[8] Difference of Gaussian
[9] Local Region
[10] Local Gradient Distribution

تاکنون در تحقیقات متنوعی، محققان تلاش کرده‌اند نسخه‌های بهبود یافته ای از عملگر SIFT را برای کاربردهای متنوع پردازش تصویر از جمله تشخیص شی، تشخیص چهره، بازیابی تصویر، دسته بندی بافت تصویر و غیره ارائه دهند.

به عنوان مثال، در [138]، که و سوکتانکار روشی تحت عنوان PCA-SIFT ارائه داده اند که با استفاده از الگوریتم آنالیز مولفه دستور دهنده[1] (PCA)، ابعاد عملگر SIFT را در حالت رایج از 128 بعد به 36 بعد کاهش می دهد . در تلاشی دیگر، میکولاجسزیک و اشمید در [139]، نسخه بهبود یافته‌ای از SIFT تحت عنوان GLOH به عنوان مخفف هیستوگرام مکان و جهت گرادیان[۲] ارائه دادند که مقاومت و قدرت تمایز پذیری[۳] SIFT را افزایش می‌دهد.

## ج- Bi-dimensional Emprical Mode Decomposition

تجزیه مدل تجربی (EMD) [142]، یکی از روش های آنالیز بافت تصویر است که در زمره روش های چند وضوحی[۴] قرار می گیرد. به صورت تئوری، EMD یک سیگنال ثابت یک بعدی را نشان می دهد که معرف مجموع مولفه های مدولاسیون جریان و مدولاسیون فرکانس می باشد. EMD در حوزه پردازش سیگنال معمولاً برای آنالیز سیگنال های یک بعدی غیر خطی کاربرد دارد. عملگر EMD، سیگنال ورودی یک بعدی را به یک سری سیگنال های باند محدود که توابع نمای ذاتی[۵] (IMF) نامیده می شوند تجزیه[۶] می‌نماید. نونز و همکارانش در [143]، نسخه دو بعدی EMD را تحت عنوان تجزیه مدل تجربی دو بعدی (BEMD) ارائه می دهند که سیگنال دو بعدی را به توابع نمای ذاتی دو بعدی (2D-IMF) تجزیه می‌نماید. به دلیل ساختار دو بعدی، BEMD در بحث پردازش تصویر، کاربرد وسیعی جهت آنالیز بافت تصویر دارد. مراحل استخراج ویژگی توسط عملگر BEMD را می توان به صورت خلاصه این چنین بیان کرد :

- مشخص کردن مکان های ماکزیمم و مینیمم تصویر توسط عملگرهای مورفولوژیک و عملگرهای کوتاه ترین فاصله[۷]
- درون یابی سطح بین تمام نقاط ماکزیمم و مینیمم توسط الگوریتم RBF و تولید دو محدوده $X_{max}$ و $X_{min}$
- میانگین گیری بین دو محدوده تعیین شده و تولید محدوده $X_M$
- تفریق تصویر از محدوده میانگین : $h_1 = I - X_M$

[1] Principal Component Analysis (PCA)
[2] Gradient Location and Orientation Histogram (GLOH)
[3] Distinction
[4] Multiresolution
[5] Intrinsic Mode Function (IMF)
[6] Decompose
[7] Geodesic operator

❖ تکرار مراحل فوق تا زمانی که $h_1$ تولید شده شرایط یک IMF دو بعدی را داشته باشد

### 2-1-3- روش های مبتنی بر مدل

بسیاری از روش‌های آنالیز بافت تصویر را تحت عنوان روش های مبتنی بر مدل می نامیم. این روش ها، تلاش می کنند به کمک ساختارهای پیش ساخته ای، الگوی تکرار در تصویر یا بافت تصویر را مدل کنند. به عنوان مثال در [54] برای مدل کردن پارچه های سالم از میدان گوسی تصادفی مارکف[1] استفاده شده است. از جمله دیگر روش های این حوزه می توان به مدل های فرکتال[2] [55و 56]، ملاک بافت[3] و مدل پسرفت خودکار[4][57] اشاره کرد.

### 2-1-4- روش های مبتنی بر فیلتر

برخی از روش ها برای آنالیز بافت تصویر از فیلترهایی از پیش طراحی شده، استفاده می کنند . این دسته از روش ها، ابتدا فیلتر را طراحی کرده و سپس تصویر مورد نظر را از فیلتر عبور داده و خروجی مورد نظر را از دیدگاه های متفاوت آنالیز می کنند . عموماً فیلترها در یکی از دو حوزه زیر تعریف می شوند:

الف. حوزه فضایی[5][40 و 41] ب. حوزه فرکانس[6][42 و 43]

برخی فیلترها نیز به صورت ترکیبی و مختلط در هر دو حوزه تعریف می‌شوند .[44، 45، 46]

**الف- حوزه فضایی**

در علم پردازش تصویر، حوزه فضایی در واقع به دامنه پیکسل ها و شدت روشنایی آنها گفته می شود. بنابراین فیلترهایی که در این حوزه تعریف می شوند، فیلترهایی هستند که به صورت مستقیم بر روی شدت روشنایی پیکسل ها و مکان هندسی آنها در تصویر اعمال می شوند به عنوان مثال، همانطور که در بخش (2-1-2-1) عنوان شد، فیلترهایی خاصی برای تشخیص لبه در تصاویر طراحی شده اند، اگر به عملکرد این فیلترها دقت کنیم، آنها برای تشخیص لبه، شدت روشنایی نقاط همسایه هر پیکسل را با آن می سنجند. بنابراین حوزه عملکرد آنها به طور مستقیم بر روی مکان فیزیکی و شدت روشنایی نقاط است . از این رو است که فیلترهای لبه یابی را می توان در زمره روش های آنالیز بافت تصویر مبتنی بر فیلترهای حوزه فضایی نیز قرار

---
[1] Gaussian Markov Random Field
[2] Fractal Model
[3] Exemplar
[4] Autoregressive
[5] Spatial Domain
[6] Frequency Domain

داد. به عنوان مثال در[47]، کلونس و همکارانش، الگوریتمی برای آنالیز بافت تصاویر بر اساس ترکیبی از نتایج فیل‌تر کردن فرکانسی همسو [1]، فیلتر کردن طیف ها [2] و فیلتر کردن محیط فضایی (دامنه) ارائه می‌دهند. کلونس و همکارانش برای فیلتر کردن تصویر در محیط فضایی و استخراج ویژگی های بافتی تصویر از فیلترهای مشهور لبه یابی استفاده نموده اند. از دیگر انواع فیلترهایی که برای آنالیز بافت تصویر استفاده می شوند می توان به فیلتر تابع گوسی [3][48]، فیلتر لاپلاسین [49]، فیلترهای پاسخ ضربه محدود خطی [4][50] و ... اشاره کرد که توضیح هر کدام مفصل بوده و در حوزه این مقاله نمی باشد.

**ب- حوزه فرکانس**

برخی از فیلترها در حوزه فرکانس تعریف شده و در آن محیط بر روی تصویر اعمال می شوند. بنابراین، برای اینکه بتوان از آنها در رابطه با تصاویر دیجیتال استفاده کرد، می بایست باید ابتدا تصویر مورد نظر توسط تبدیلات موجک به حوزه فرکانس رفته تا تعریف سیگنالی آن حاصل شود. سپس فیلترهای مورد نظر بر روی مولفه های فرکانسی آن اعمال گردد. از جمله فیلترهای این حوزه می توان به تبدیل موجک [5][51]، فیلترهای گابور [52] و فیلترهای قلم[6][53] اشاره کرد. اطلاعات استخراجی از محیط فرکانس معمولاً نسبت به ویژگی های محیط فضایی پایدارتر بوده و از این جهت کاربرد وسیعی نیز دارند . ولیکن بار محاسباتی بالا و زمان اجرای بیشتر از جمله معایب برخی از این گروه فیلترها می باشد.

## 2.2. Texture Classification Methods

همانطور که در فصل قبل نیز توضیح داده شد، دسته‌بندی[7] به معنای نسبت دادن یک نمونه به یکی از گروه های از پیش تعریف شده می باشد. در دسته‌بندی بافت تصویر [8] نیز هدف نسبت دادن بافت یک تصویر نمونه ی آزمایشی به یکی از کلاس های بافتی از پیش تعریف شده است . روش های ارائه شده برای دسته بندی بافت تصویر معمولاً شامل دو مرحله اصلی : الف. مرحله آموزش[9] (استخراج ویژگی) و ب. مرحله تشخیص [10] (دسته بندی) [7، 8، 9] می باشند.

---

[1] Isotropic Frequency Filtering
[2] Spectral Filtering
[3] Gaussian Function Filter
[4] Linear Finite Impulse Response Filter
[5] Wavelet Filter
[6] Wedge Filter
[7] Classification
[8] Image Texture Classification
[9] Learning Phase
[10] Recognition Phase

در مرحله آموزش، هدف ساختن یک مدل برای هر کدام از بافت هایی است که در تصاویر آموزشی پایگاه[1] وجود دارد. بنابراین، تمامی روش‌هایی که پیش‌تر برای آنالیز بافت تصویر مطرح شد، می‌توانند در این مرحله مورد استفاده قرار گیرند. به همین منظور، کافی است که خروج ی آنالیز بافت به صورت هایی همچون ویژگی های عددی، هیستوگرام های گسسته، توزیع های تجربی[2]، خصایص آماری، خصایص بافتی [3] همچون تباین، ساختار فضایی، جهت و .... تبدیل گردد تا امکان مقایسه و دسته بندی آنها وجود داشته باشد . به عبارت ساده تر در این مرحله می‌بایست یک سری ویژگی تفکیک کننده از بافت تصویر استخراج گردد . از آنجایی که هدف اصلی این مرحله، آموزش بافت های مذکور به سیستم است، برخی محققان آن را مرحله یادگیری[4] نیز می‌نامند.

در مرحله دوم، با استفاده از انواع دسته بند ها می بایست کلاس بافتی تصویر مورد آزمایش مشخص گردد. در این مرحله نوآوری کمتر بوده زیرا در اکثر موارد از دسته بندهای متعارفی همچون انواع شبکه های عصبی، انواع دسته بندهای خطی مانند آنالیز تفکیک کننده خطی [5] (LDA)، دسته بندهای آماری همچون نزدیک ترین همسایه [6] (KNN)، ماشین بردار پشتیبان[7](SVM) و ....، استفاده می گردد. البته در برخی از تحقیقات نیز محققان تلاش می - کنند که با تغییر پارامترهای ورودی، دسته بندهای کلاسیک را با مسئله و کاربرد مورد نظر تحقیق سازگار[8] کنند.

دسته وسیعی از محققان در این حوزه به این نظریه اعتقاد دارند که "در شناسایی و درک تصاویر، بافت مستقل از دیگر ویژگی های تصویر از جمله رنگ و شکل [9] آن است". به همین منظور، تاکنون روش‌های بسیار متنوعی برای آنالیز بافت تصاویر با سطوح خاکستری [10] ارائه شده است. نویسنده این مقاله با این نظریه موافق نبوده و اعتقاد دارد که "**علی الرغم استقلال** بافت تصویر از رنگ و شکل ، ولیکن در درک و دسته‌بندی تصاویر، ترکیب ویژگی های بافت و رنگ می تواند دقت دسته بندی [11] را افزایش دهد ". در زیر بخش 2-4 به این نظریه بیشتر

1 Training Data
2 Empirical Distribution
3 Texture Properties
4 Learning Phase
5 Linear Discriminant Analysis (LDA)
6 K Nearest Neighbor (KNN)
7 Support Vector Machine (SVM)
8 Adapt
9 Shape
10 Gray-Level Texture Images
11 Classification Accuracy

پرداخته خواهد شد. ولی فعلاً در زیر بخش 2-3، برخی از روش‌های به روز [1] و پرکاربرد برای آنالیز بافت تصاویر با سطوح خاکستری به اختصار بررسی و نقد خواهد شد.

## 2.3. Review on Gray-level Texture Classification

استفاده از ویژگی های آماری یکی از قدیمی ترین راهکارها برای آنالیز انواع تصاویر بخصوص تصاویر بافتی می باشد. به عنوان مثال د ر سال 1995، کویی‌چن و همکارانش [58] ، روشی برای دسته بندی بافت تصاویر با سطوح خاکستری بر پایه ویژگی های آماری هندسی [2] ارائه کردند. ویژگی‌هایی همچون نامنظمی [3]، تراکم [4]، احاطگی [5]، میانگین شدت روشنایی و انحراف از استاندارد [6] در این روش برای آنالیز بافت تصویر مورد ا ستفاده قرار گرفته اند. کویی‌چن و همکارانش در [58]، روش ارائه شده خودشان را با سه روش ماتریس وابستگی فضایی سطوح خاکستری [7] (SGLDM)، ویژگی های لیو بر اساس تبدیل فوریه و ماتریس ویژگی های آماری [8] (SFM) مقایسه کرده و نشان می‌دهند که روش ایشان دقت دسته بندی بالاتری را فراهم می کند. در نقد این مقاله می توان دو مزیت و سه عیب را ذکر کرد که در جدول 2-1 آورده شده است.

**جدول 2-1. تحلیل مزیت ها و عیوب دسته بندی بر اساس ویژگی های آماری هندسی [58]**

| مزایا | معایب |
|---|---|
| • بار محاسباتی ویژگی های آماری به دلیل فرمول مشخص، کم است | • دقت دسته بندی ارائه شده پایین است<br>• ویژگی های آماری نسبت به نویز حساس هستند<br>• روش ارائه شده را نمی توان به تصاویر رنگی تعمیم داد<br>• ویژگی های آماری ذاتاً نسبت به چرخش تصویر حساس هستند |

مزایای استفاده از ویژگی های آماری باعث شده است که در طی سالیان اخیر همچنان از این گروه ویژگی‌ها، خواه به صورت مستقل و خواه به صورت ترکیب با دیگر عملگرهای آنالیزی بافت، جهت دسته بندی بافت تصویر استفاده گردد. به عنوان مثال در [59]، وارما و زیسرمن

1 State-Of-the-art
2 Geometrical Statistical Features
3 Irregularity
4 Compactness
5 Perimeter
6 Standard Deviation
7 Spatial Gray Level Dependence Matrix (SGLDM)
8 Statistical Features Matrix

الگوریتمی برای دسته بندی بافت بر پایه استخراج ویژگی های آماری ارائه داده اند که فقط از یک نمونه تصویر آموزشی استفاده می کند. در [59]، تصویر آموزشی بر اساس توزیع احتمالی پاسخ فیلتر ضربه مدل شده و سپس هیستوگرام آن رسم گردیده و در نهایت ویژگی آماری از هیستوگرام خروجی استخراج می گردد. در [59]، از تابع فاصله مبتنی بر معیار $x^2$، برای مرحله دسته بندی استفاده شده است. مزایا و معایب این روش در جدول 2-2 بیان شده است.

**جدول 2-2. تحلیل مزیت ها و عیوب دسته بندی بر اساس ویژگی های آماری از تصویر تنها [59]**

| مزایا | معایب |
|---|---|
| • حساسیت اندک نسبت به نقطه نظر تصویر برداری و روشنایی نامتعارف در تصویر<br>• عدم حساسیت به چرخش در تصویر | • بار محاسباتی بالا به دلیل چند مرحله اعمال فیلتر و استخراج هیستوگرام<br>• دقت دسته بندی بالایی فراهم نمی کند<br>• روش ارائه شده را نمی توان به تصاویر رنگی تعمیم داد |

در تحقیقات متنوعی، برخی از محققان تلاش کرده اند که از ترکیب ویژگی‌های آماری با دیگر عملگرهای آنالیزی، ویژگی‌هایی با قدرت تفکیک پذیری بالاتر فراهم کنند . به عنوان مثال در [60]، از ترکیب ویژگی های آماری و ویژگی های موجک [1] جهت دسته بندی بافت تصاویر تومورهای مغزی انسان استفاده شده است. در [60]، پس از اعمال تابع تبدیل موجک بر روی تصویر، ویژگی های آماری همچون بی نظمی [2]، تباین [3]، تجانس [4]، همبستگی [5] و لحظه دوم زاویه‌ای [6] استخراج می گردد. در این مقاله از دسته‌بندهای ماشین بردار پشتیبان و نزدیک ترین همسایه برای محاسبه دقت دسته‌بندی [7] استفاده گردیده است. این روش علی الرغم دقت دسته بندی بالا که بر روی پایگاه تصاویر MRI مغز فراهم می نماید، معایبی نیز دارد که در جدول 2-3، بیان شده است.

[1] Wavelet Features
[2] Entropy
[3] Contrast
[4] Homogeneity
[5] Correlation
[6] Angular Second Moment
[7] Classification Accuracy

**جدول 2-3. تحلیل مزیت ها و عیوب دسته بندی بر اساس ویژگی های آماری مرتبه اول و دوم در تبدیل موجک [60]**

| مزایا | معایب |
|---|---|
| • دقت دسته بندی بالا بر روی تصاویر MRI مغز انسان<br>• بار محاسباتی نسبتا پایین در مقایسه با روش های مطرح شده در حوزه تصاویر MRI<br>• سازگاری با انواع دسته‌بند ها | • عدم سازگاری احتمالی با تصاویر بافتی غیر از مغز انسان<br>• حساسیت نسبت به چرخش، نویز، روشنایی[1]<br>• عدم امکان تعمیم روش به تصاویر رنگی به دلیل استفاده از تابع تبدیل موجک |

در [63]، بارلی و توون الگوریتمی برای دسته بندی بافت تصویر بر اساس ترکیب ویژگی های تبدیل موجک، ویژگی‌های گابور موجک، هرم هدایت شونده[2] و ماتریس‌های همرخدادی سطوح خاکستری ارائه کرده اند . روش ترکیبی خلاقانه ارائه شده در [63]، تلاش دارد که از مزایای تمامی عملگرهای فوق یکجا استفاده نماید . این امر نارسایی هایی را هم نتیجه داده شده است که در جدول زیر نشان داده شده است.

**جدول 2-4. تحلیل مزیت ها و عیوب دسته بندی بر اساس ترکیب تبدیل موجک، موجک گابور و هرم هدایت شده [63]**

| مزایا | معایب |
|---|---|
| • کارایی بالا درمواجهه با انواع بافت ها<br>• عدم حساسیت به تغییر مقیاس تصویر<br>• ارائه دقت دسته بندی بالا با استفاده از انواع دسته بندهای کلاسیک | • بار محاسباتی بسیار زیاد به دلیل ترکیب 5 گروه عملگر آنالیزی<br>• نیاز به تغییرات بنیادین برای تعمیم روش به تصاویر رنگی و در نتیجه بار محاسباتی بالاتر |

ویژگی‌های تبدیل موجک که در مثال قبل نیز مطرح شد، از دیرباز مورد علاقه محققین جهت آنالیز و دسته بندی بافت تصاویر بوده است . در یکی از تحقیقات قدیمی [61] که در سال 1993، انجام شد از ویژگی های حاصل از اعمال بانک فیلتر موجک برای دسته بندی بافت تصاویر در یک پایگاه دست ساز شامل 25 تصویر طبیعی استفاده شد . در [61] برای مرحله آزمایش از دسته بند شبکه عصبی دولایه ساده [3] استفاده شده است . مزایای مثبت استفاده از تابع تبدیل موجک باعث گردیده است که محققین در سالهای اخیر بر روی ترکیب این ویژگی-ها با انواع دیگر عملگر های آنالیز بافت تمرکز نمایند . به عنوان مثال در [62]، روشی بر پایه ترکیب خروجی تبدیل موجک و ویژگی های ماتریس‌های همرخدادی برای دسته بندی بافت

[1] Illumination
[2] Steerable Pyramid
[3] Simple Two-Layer Neural Network

تصاویر با سطوح خاکستری ارائه شده است. در این روش، ابتدا تصویر تحت تابع تبدیل موجک به فضای فرکانس منتقل شده و سپس از تصویر تبدیل یافته، ماتریس‌های همرخدادی در جهات گوناگون استخراج می گردد. نویسندگان این مقاله مهم ترین دستاورد خودشان را ارائه روشی مقاوم به تغییر مقیاس[1] (زوم شدن) تصویر اعلام می نمایند. جدول 2-4 مزایای و معایب این روش را از دیدگاه نویسنده مقاله بیان می کند.

**جدول 2-5. تحلیل مزیت ها و عیوب دسته بندی بر اساس ترکیب ماتریس همرخدادی و تبدیل موجک [62]**

| مزایا | معایب |
| --- | --- |
| • دقت دسته بندی مناسب<br>• عدم حساسیت به تغییر مقیاس تصویر | • حساسیت نسبت به چرخش و نویز به دلیل استفاده از ماتریس‌های همرخدادی<br>• عدم امکان تعمیم روش به تصاویر رنگی به دلیل استفاده از تابع تبدیل موجک |

ترکیب ویژگی‌های استخراجی از فیلتر موجک و فیلتر گابور نیز در [64] برای دسته بندی بافت در تصاویر دریافت شده از رادار مورد بررسی قرار گرفته است. بزرگترین مزیت آن را می-توان عدم حساسیت به چرخش در کنار دقت دسته بندی مناسب برای کاربرد مطرح شده عنوان کرد. فارغ از ویژگی های آماری و تبدیلات محیط فرکانس، روش های متنوع دیگری نیز برای دسته‌بندی بلفت تصاویر با سطوح خاکستری مطرح شده اند که از آن جمله می توان به گروه ویژگی‌های ساختاری اشاره کرد. یکی از معروف ترین عملگرهای آنالیز ساختاری، الگوی دودویی محلی[2] (LBP) است. الگوهای دودویی محلی یک عملگر آنالیزی است که ساختار فضایی محلی[3] و تباین محلی[4] تصویر را تعریف می نماید. این عملگر اولین بار توسط پیتکانن و اوجالا در [65]، برای آنالیز بافت تصویر ارائه شد. عدم حساسیت نسبت به چرخش و بار محاسباتی مناسب، محققین را بر آن داشت تا در سالهای بعد نسخه های بهبود یافته آن را ارائه دهند. از آنجایی که در این مقاله نیز از همین عملگر برای دسته بندی بافت تصاویر رنگی استفاده شده است، توضیح جزییات آن را به فصل سوم موکول کرده و در ادامه تنها به مرور مقالاتی می پردازیم که از این عملگر جهت دسته بندی بافت تصویر استفاده کرده‌اند. در [66]، اوجالا و همکارانش، نسخه بهبود یافته‌ای از الگوهای دودویی محلی را ارائه کرده و از ترکیب آن

[1] Multi-Scale
[2] Local Binary Patterns
[3] Local Spatial Structure
[4] Local Contrast

با ویژگی‌های تباین برای دسته بندی بافت تصاویر استفاده کرده اند. یکی از خلاقیت های این مقاله، استفاده از معیار عدم شباهت لگاریتم درست نمایی برای مرحله دسته‌بندی است. مزایای فوق‌العاده این الگوریتم باعث شده است که این مقاله یکی از پر استناد ترین[۱] مقالات در حوزه آنالیز بافت تصویر باشد . در جدول زیر، مزایا و معایب این الگوریتم از دیدگاه نویسنده مورد بررسی قرار گرفته است.

**جدول 2-6. تحلیل مزیت ها و عیوب دسته بندی بر اساس نسخه مستقل از چرخش الگوی دودویی محلی** [**66**]

| مزایا | معایب |
|---|---|
| • دقت دسته‌بندی مناسب<br>• عدم حساست به چرخش[۲]<br>• حساسیت کم به تغییر مقیاس تصویر[۳] | • حساسیت به نویز به ویژه نویز ضربه ای<br>• عدم تعریف اولیه برای تصاویر رنگی |

قریب به ده سال پس از مقاله [66]، لیو و همکارانش در [67]، نسخه دیگری از الگوهای دودیی محلی برای دسته بندی بافت تصاویر ارائه کردند . در [67]، چهار عملگر جدید به نام های شدت روشنایی مرکزی الگوهای دودویی محلی (CI-LBP)[۴]، شدت روشنایی همسایگان الگوهای دودویی محلی[۵] (NI-LBP)، تفاوت محوری الگوهای دودویی محلی[۶] (RD-LBP) و تفاوت زاویه‌ای الگوهای دودویی محلی[۷] (AD-LBP) ارائه می گردد که ترکیب آنان دقت دسته بندی را افزایش داده است. بزرگترین نقطه ضعف [67] را می توان بار محاسباتی بسیار بالا به دلیل پیاده سازی چهار عملگر مجزا و عدم استفاده از ویژگی های رنگ عنوان کرد

ویژگی های استخراجی از فیلترهای لبه یابی نیز در زمره گروه وی ژگی‌های ساختاری قرار می گیرند. ویژگی‌های لبه علی‌الرغم آنکه به خودی خود توانایی بالایی در دسته بندی بافت ندارند ولی در ترکیب با دیگر ویژگی‌ها می توانند نتایج مفیدی حاصل کنند. به عنوان مثال در [68]، ترکیبی نوآورانه از ویژگی های استخراجی از فیلترهای لبه یابی ، الگوهای دودویی محلی و ماتریس های همرخدادی برای دسته بافت تصاویر با سطوح خاکستری ارائه شده است . روش

[1] Hot Cited
[2] Rotation Invariant
[3] Multi Scale Invariant
[4] Center Intensity Local Binary Patterns (CI-LBPP)
[5] Neighbors Intensity Local Binary Patterns (NI-LBPP)
[6] Radial Difference Local Binary Patterns (RD-LBPP)
[7] Angular Difference Local Binary Patterns (CI-LBPP)

سه مرحله ای ارائه شده در این مقاله، دقت بالاتری نسبت به استفاده منفرد از عملگرهای آنالیزی الگوی دودویی محلی ساده و ماتریس های همرخدادی فراهم آورده است که د لیل آن را می توان ترکیب با ویژگی های لبه دانست. نقد [68] در جدول زیر آورده شده است.

**جدول 2-7. تحلیل مزیت ها و عیوب دسته بندی بر اساس ترکیب الگوی دودویی محلی و ماتریس همرخدادی و لبه [68]**

| مزایا | معایب |
|---|---|
| • دقت دسته‌بندی بالاتر نسبت به نسخه اولیه الگوی دودویی محلی<br>• عدم حساست به چرخش | • بار محاسباتی بالا به دلیل فاز آموزش سه مرحله ای<br>• حساسیت به نویز ضربه ای به دلیل استفاده از الگوی دودویی محلی<br>• عدم امکان تعمیم آسان به تصاویر رنگی |

در تحقیقی دیگر، ترکیب ویژگی های لبه با عملگرهای موفولوژیک برای دسته بندی بافت مورد بررسی قرار گرفته است . در [69]، ابتدا عملگر مورفولوژیک بازکنندگی [1] بر روی تصویر اعمال شده و سپس لبه یابی می گردد. در نهایت میزان رخداد الگوهای لبه در چهار جهت عمودی، افقی، چپ و راست به عنوان ویژگی استخراج می گردد. نویسندگان [69]، ارائه الگوریتمی کارآمد با پیاده‌سازی آسان را هدف اصلی خود بیان می‌دارند.

تاکنون الگوریتم های بسیار متنوع دیگری نیز برای دسته بندی بافت تصاویر با سطوح خاکستری ارائه شده است. در مثالی متفاوت، در [70]، فکری ارشاد، روشی بر اساس تغییرات انرژی در تصاویر فیلتر شده توسط عملگرهای الگوی دودوی ی محلی، لبه یابی و ماتریس همرخدادی ارائه کرده است. در الگوریتم [70]، ابتدا عملگر الگوی دودویی محلی بر روی تصویر اعمال شده و سپس انرژی تصویر خروجی با استفاده از معادله 2-6 محاسبه می‌گردد. در ادامه، همین داستان بر پایه فیلتر لبه یابی سوبل و ماتریس همرخدادی نیز تکرار می گردد. در نهایت میزان تفاوت انرژی تصویر اصلی با انرژی‌های محاسبه شده در هر حالت، به عنوان یکی از ابعاد بردار ویژگی نهایی در نظر گرفته می شود. سازگاری مناسب بردار ویژگی با انواع دسته بندها و دقت دسته بندی بالا از جمله محاسن [2] این الگوریتم می باشد. استفاده از اطلاعات آماری هیستوگرام نیز در موارد متعددی جهت آنالیز بافت مورد استفاده قرار گرفته که در برخی موارد نتایج مثبتی نیز به همراه داشته است . به عنوان مثال در [71]، از هیستوگرام خروجی تصویر تحت عملگر الگوی دودویی محلی برای دسته بندی بافت تصاویر با کاربرد مشخص جهت

[1] Opening
[2] Advantages

تشخیص دود[1] استفاده شده است. در [71]، ابتدا الگوی دودویی محلی ساده بر روی تصویر اعمال شده و سپس گرادیان کدهای خروجی محاسبه می گردد و در نهایت هیستوگرام هایی بر اساس نرخ رخداد گرادیان ها تشکیل می شود. در این الگوریتم، برای مرحله دسته بندی نیز از ماشین بردار پشتیبان استفاده شده است. نویسندگان این مقاله نکته مثبت روش ارائه شده را در نظر گرفتن رابطه بین کدهای الگوهای دودویی محلی اعلام می کنند. همانطور که پیش تر در بخش 2-1-2 اشاره شد، عملگر BEMD یکی از عملگرهای کارآمد در حوزه آنالیز سیگنال های دو بعدی و پردازش تصویر است. ژیاوو و همکارانش در [144]، از این عملگر جهت آنالیز تصاویر بافتی و استخراج ویژگی های بافت تصویر استفاده کرده اند. در [144]، ابتدا تصویر بافتی توسط BEMD به تعدادی توابع نمای ذاتی (IMF)شکسته شده و سپس خصایص محلی[2] IMF های دو بعدی استخراج می گردد. روش‌های ارائه شده در حوزه دسته بندی بافت، به ویژه در تصاویر با سطوح خاکستری، بسیار متنوع است. بنابراین در راستای پیشبرد اهداف این مقاله، برخی از مقالات و کتاب ها در این حوزه بر اساس فاکتورهایی نظیر اعتبار مجلات، میزان استناد، میزان ارتباط با موضوع اصلی مقاله و مزایای ارائه شده در الگوریتم، انتخاب و مورد بررسی قرار گرفت که برخی از آنها در این بخش توضیح داده شد. ولی ارائه جزییات تمامی آنها در حوصله این مقاله نمی‌گنجد. بنابراین در جدول زیر به اختصار به برخی از این روش‌ها اشاره شده است.

---
[1] Smoke Detection
[2] Local Properties

**جدول 2-8.** **بررسی برخی از روش های ارائه شده جهت دسته بندی وآنالیز بافت تصاویر با سطوح خاکستری**

| ردیف | سال ارائه | روش ارائه شده | مرجع |
|---|---|---|---|
| 1 | 1992 | مدل کردن پسرفت خودکار[1] چند مقیاسی | [57] |
| 2 | 1993 | بانک ویژگی‌های تبدیل موجک | [61] |
| 3 | 1995 | ویژگی‌های آماری هندسی | [58] |
| 4 | 1995 | ویژگی‌های استخراجی از قاب‌های موجک[2] | [72] |
| 5 | 1999 | مدل کردن ویژگی‌های گابور-موجک در محیط ترکیبی فرکانس و فضا[3] | [73] |
| 6 | 1999 | آنالیز فرکتال دو بعدی | [74] |
| 7 | 2001 | تبدیل فوریه محلی | [75] |
| 8 | 2001 | آنالیز مولفه‌های دستوری تحت کرنل[4] | [76] |
| 9 | 2002 | نسخه مستقل از چرخش و چند مقیاسی الگوهای دودویی محلی | [66] |
| 10 | 2002 | ترکیب ویژگی‌های آماری و ماتریس‌های همرخدادی مستخرج از تبدیل موجک | [62] |
| 11 | 2003 | تاثیر انرژی قطبی موجک[5] | [77] |
| 12 | 2003 | ویژگی های هیستوگرام طیفی[6] | [78] |
| 13 | 2004 | تبدیل ویژگی مستقل از مقیاس (SIFT) | [137] |
| 14 | 2005 | ترکیب تبدیل ویژگی های مستقل از مقیاس (SIFT) و الگوریتم آنالیز مولفه دستور دهنده[7] (PCA) | [138] |
| 15 | 2006 | ترکیب ویژگی‌های مرتبه اول و دوم آماری و ماتریس‌های همرخدادی وفرکتال | [79] |
| 16 | 2007 | مدل ترکیب گوسی[8] به همراه بانک فیلترهای موجک | [80] |
| 17 | 2007 | نسخه بهبود یافته الگوهای دودویی محلی به همراه توزیع فضایی نمونه های غالب[9] | [81] |
| 18 | 2007 | تراکم گوسی تعمیم یافته[10] برای مدل کردن فرکانس‌های بالا موجک | [48] |
| 19 | 2008 | آنالیز میزان همبستگی به کمک مدل پسرفت خطی[11] بر اساس تبدیل موجک | [82] |
| 20 | 2008 | بخش‌بندی فاز محلی مستخرج از چهار ضریب فرکانس پایین | [83] |
| 21 | 2008 | الگوهای بافتی محلی [12] | [84] |
| 22 | 2009 | الگوهای دودویی محلی غالب | [85] |
| 23 | 2009 | ترکیب تجزیه مدل تجربی دو بعدی (BEMD) و عملگر آنالیز محلی ساختار چهار وجهی[13] | [144] |
| 24 | 2010 | فیلترهای گابور انتخابی توسط روش‌های انتخاب ویژگی[14] | [86] |

---

[1] Autoregressive
[2] Wavelet Frames
[3] Spatial-Frequency Domain
[4] Kernel Principal Component Analysis (KPCA)
[5] Log-Polar Wavelet Energy Signature
[6] Spectral Histogram
[7] Principal Component Analysis (PCA)
[8] Gaussian Mixture Model(GMM)
[9] Spatial Distribution of Dominant Patterns
[10] Generalized Gaussian Density (GDD)
[11] Linear Regression Model
[12] Local Texture Patterns (LTP)
[13] Quaternionic Representation
[14] Feature Selection

| | | | |
|---|---|---|---|
| [87] | ارائه روشی چند مرحله ای در حد پیکسل [1] بر اساس الگوی انتخاب خروجی فیلترها | 2010 | 25 |
| [68] | ترکیب ویژگی های استخراجی لبه، الگوهای دودویی محلی و ماتریس های همرخدادی | 2011 | 26 |
| [88] | نگاشت مرتب شده جهت ترکیب خروجی عملگرهای مختلف آنالیز بافت | 2012 | 27 |
| [60] | ترکیب ویژگی های مرتبه اول و دوم همچون ویژگی های آماری و ویژگی های موجک | 2012 | 28 |
| [67] | نسخه بهبود یافت الگوهای دودویی محلی بر اساس شدت روشنایی مرکزی، شدت روشنایی همسایگان، تفاوت محوری و تفاوت زاویه‌ای | 2012 | 29 |
| [70] | تغییرات انرژی در خروجی های ماتریس‌های همرخدادی | 2012 | 30 |
| [89] | ارائه روشی برای انتخاب تصادفی ویژگی های بافتی بر اساس روش کیف کلمات [2] | 2012 | 31 |
| [90] | نسخه مستقل از نویز الگوهای دودویی محلی | 2013 | 32 |
| [91] | ارائه نوعی از ویژگی بی‌نظمی احتمالی مبتنی بر اطلاعات فیلتر گوسین | 2013 | 33 |
| [92] | الگوهای پشتیبان محلی بر پایه الگوهای دودویی محلی | 2014 | 34 |
| [63] | ترکیب عملگرهای ماتریس های همرخدادی، گابور موجک و هرم هدایت شونده | 2014 | 35 |
| [93] | ارائه عملگری بر مبنای جهت محلی [3] پیکسل ها | 2015 | 36 |
| [94] | ترکیب الگوهای دودویی میانه [4] بر اساس تغییر مقیاس همسایگی | 2015 | 37 |
| [95] | ترکیب هیستوگرام های تولیدی پله‌ای [5] در زیر باندهای تبدیل موجک [6] | 2015 | 38 |
| [96] | ارائه روشی بر پایه برنامه ریزی ژنتیک [7] برای انتخاب سازگارترین دسته ویژگی - های بافت | 2015 | 39 |
| [97] | نسخه بهبود یافته الگوهای دودویی محلی بر اساس میانه شدت روشنایی محلی | 2016 | 40 |
| [98] | ویژگی‌های تفاوت‌های بخش‌های کوچک متراکم [8] در تصویر | 2016 | 41 |
| [99] | نسخه بهبود یافته مقاوم در برابر نویز الگوهای دودویی محلی [9] | 2016 | 42 |

## 2.4. Color Texture Classification

چشمان انسان زمانی که یک تصویر را مشاهده می نماید، آن را به صورت ترکیبی از مولفه هایی همچون بافت، شکل [10] و رنگ درک می کند[10]. بخش عمده‌ای از روش‌هایی که تاکنون برای آنالیز بافت تصویر ارائه شده‌اند، فقط بر روی تصاویری با سطوح خاکستری [11] عمل می‌کنند. در حالی که یکی ا ز ویژگی‌های کلیدی که می تواند در دسته بندی و آنالیز تصاویر استفاده شود،

[1] Multi Level Pixel-Based
[2] Bag of Words
[3] Local Orientation Adaptive
[4] Median Binary Pattern (MBP)
[5] Heterogeneous and Incrementally Generated Histograms (HIGH)
[6] Wavelet Sub bands
[7] Genetic Programming
[8] Dense Micro-Block Differences (DMD)
[9] Noise-Tolerant Local Binary Patterns (NTLBP)
[10] Shape
[11] Gray Levels Images

رنگ است. ترکیب ویژگی‌های رنگ تصویر در کنار ویژگی هایی همچون شکل و حرکت در کاربردهای گوناگونی از حوزه پردازش تصویر استفاده شده و نتایج مثبتی را به ارمغان آورده است که از آن جمله می توان به تشخیص اشیا[1] [11] و ردیابی حرکت[2] [12] اشاره کرد. بخش عمده‌ای از تصاویری که ما با آنها سر و کار داریم، رنگی هستند، بنابراین در نظر نگرفتن ویژگی‌های رنگ، در آنالیز چنین تصاویری به هیچ عنوان معقول نیست . اهمیت این موضوع و همچنین تحقیقات نه چندان وسیعی که تاکنون در این حوزه صورت گرفته، ما را بر آن داشت تا در این مقاله، راهکاری برای دسته‌بندی بافت تصاویر رنگی ارائه نمایم . در راستای این هدف اصلی، در این مقاله تمرکز اصلی بر روی مرحله یادگیری بوده و تلاش می گردد که عملگری برای آنالیز تصاویر رنگی ارائه گردد که ویژگی‌های استخراجی از بافت و رنگ را با یکدیگر ترکیب نماید تا ماکزیمم دقت دسته بندی و ماکزیمم توانایی تفکیک پذیری [3] بدست آید. البته در راستای افزایش دقت دسته بندی، بر روی مرحله دسته بندی نیز کارهایی صورت خواهد گرفت و تلاش می‌شود تا دسته‌بندهای کلاسیک را با ب ردارهای ویژگی استخراجی سازگار نماییم. بار محاسباتی کم، مقاومت نسبت به نویز، مقاومت نسبت به چرخش از دیگر نکاتی است که در ارائه عملگر آنالیز تصاویر بافتی، مورد نظر قرار داده شده است.

## 2.5. Review on Color Texture Classifiction Methods

در راستای ترکیبِ ویژگی های بافت و رنگ، معمولاً دو راهکار متداول وجود دارد که بیشتر از طرف محققان دنبال می شود: ترکیبِ اولیه[4] و ترکیبِ ثانویه[5] [13].

هر تصویر رنگی معمولاً نتیجه ترکیب اطلاعات در سه کانال رنگی مجزا است که به هر کدام از کانال‌ها می‌توان به چشم یک تصویر تک رنگ (تصویر دارای سطوح خاکستری) نیز نگاه کرد. در راهکار ترکیب اولیه، ابتدا اطلاعات کانال های رنگی از یکدیگر مجزا شده و سپس عملگرهای آنالیز بافت بر روی هر کانال رنگی تصویر به صورت مجزا اعمال می شوند. در نهایت یک سری ویژگی از هر کانال استخراج شده و با یکدیگر ترکیب (متصل) می‌گردند. این دسته از ویژگی-

[1] Object Recognition
[2] Motion Tracking
[3] Discriminative
[4] Early Fusion
[5] Late Fusion

های استخراجی زمانی که در تصویر ویژگی های بافت و رنگ در سطح پیکسل ترکیب شده باشند، قدرت جداسازی[1] بالایی را فراهم کرده و دقت دسته بندی را افزایش می‌دهند.
در نقطه مقابل، در راهکار ترکیب ثانویه، ویژگی های بافت و رنگ به صو رت مجزا (معمولا به شکل هیستوگرام) استخراج شده و در نهایت هیستوگرام های مجزا به یکدیگر پیوست شده [2] و بازنمایی[3] نهایی را تشکیل می‌دهند [14]. به عبارت دیگر در ترکیب ثانویه، ویژگی‌های بافت و رنگ در سطح کل تصویر محاسبه شده و در نهایت با یکدیگر ترکیب می شوند
تاکنون روش های متنوعی برای دسته‌بندی بافت تصاویر رنگی بر اساس یکی از دو راهکار فوق ارائه شده است. روش‌های هر گروه نیز در مراحل یادگیری (استخراج ویژگی) یا دسته بندی با یکدیگر تفاوت می کنند. همانطور که پیش تر اشاره شد، تمرکز اصلی این مقاله بر روی مرحله یادگیری است. بنابراین در راستای ارائه عملگر ترکیبی آنالیز بافت- رنگ، بسیاری از روش های پیشین در این حوزه مورد مطالعه قرار گرفت و مزایا و معایب هر کدام بررسی شد . در ادامه برخی از روش های کارآمد مطالعه شده به اختصار توضیح داده خواهند شد

## 2-5-1- ترکیب بافت و رنگ بر اساس نسخه بهبود یافته ماتریس های همرخدادی[4]

بنکو و همکارانش در [10]، روشی برای دسته‌بندی بافت در تصاویر رنگی بر پایه ماتریس های همرخدادی ارائه دادند. این الگوریتم در مرحله یادگیری هر تصویر رنگی را به سه کانال رنگی مجزا تقسیم کرده و سه تصویر تک رنگ می سازد. سپس ماتریس‌های همرخدادی را برای هر تصویر در 4 جهت مختلف شامل صفر، 45، 90 و 135 درجه محاسبه و تشکیل می دهد. ماتریس‌های همرخدادی از جمله عملگرهای آنالیز احتمالی هستند، بنابراین انواع ویژگی های آماری[5] همچون میانگین، واریانس، بی‌نظمی[6]، انرژی، اطلاعات مشترک[7]، انحراف از استاندارد[8] و غیره از آن قابل استخراج می‌باشد. ویژگی‌های آماری فوق را غالباً ویژگی‌های هارالیک[9] نیز می‌نامند[100]. در این مقاله نویسندگان ادعا کرده اند که بسیاری از ویژگی های آماری فوق، با

[1] Discriminative
[2] Concatenating
[3] Representation
[4] Co-occurrence matrixes
[5] Statistical Features
[6] Entropy
[7] Mutual Information
[8] Standard Deviation
[9] Haralick Features

یکدیگر همپوشانی اطلاعاتی زیادی دارند . بنابراین در جهت کاهش بار محاسباتی برای هر ماتریس ویژگی تجانس[1] را محاسبه کرده‌اند. ویژگی تجانس در معادله زیر نشان داده شده است. است.

$$Homogeneity_{d,\theta} = \sum_i \sum_j \frac{P_{d,\theta}(i,j)}{1+|i-j|^2} \qquad (11\text{-}2)$$

در معادله فوق، i و j نشان‌دهنده مختصات سطر و ستون در ماتریس همرخدادی بوده و P نیز نشان دهنده مقدار احتمالی خانه مورد نظر در ماتریس است . $\theta$ نیز جهت (زاویه) محاسبه ماتریس همرخدادی و d نیز فاصله در نظر گرفته شده در تعریف همسایگی ماتریس همرخدادی را نشان می‌دهد. بنابراین در این الگوریتم، پس از محاسبه تجانس برای هر ماتریس همرخدادی، در نهایت یک بردار ویژگی 12 بعدی به عنوان خروجی مرحله یادگیری بدست می آید. در [10]، برای مرحله دسته‌بندی از ماشین بردار پشتیبان[2] استفاده شده است. در مرحله نتیجه‌گیری، دقت دسته‌بندی روش ارائه شده، بر روی دو پایگاه داده معروف در این حوزه به نام‌های Outex و Vistex آزمایش شده است. نویسندگان دقت دسته‌بندی را در فضاهای رنگی RGB[3] و HSV[4] و همچنین تصویر تک رنگ با سطوح خاکستری بررسی کرده اند. درنهایت بالاترین دقت در پایگاه Outex به میزان 92.42 در فضای رنگی RGB و در پایگاه Vistex به میزان 90.97 در فضای HSV حاصل شد. همچنین در مقام مقایسه، نویسندگان نشان داده‌اند که استخراج ویژگی با استفاده از فیلترهای گابور برای هر کانال رنگی دقت دسته بندی چندان بالایی را به ارمغان نمی‌آورد و در بالاترین مقدار به 89.04 درصد می‌رسد. با توضیحات فوق، این روش ارائه شده را می توان در زمره‌ی راهکارهای ترکیب اولیه در نظر گرفت . مزایا و معایب این روش در جدول زیر ذکر شده است.

**جدول 2-9. تحلیل مزیت ها و عیوب دسته بندی تصاویر رنگی بر اساس نسخه بهبود یافته ماتریس های همرخدادی [10]**

| مزایا | معایب |
|---|---|
| • دقت دسته‌بندی نسبتاً بالا<br>• امکان سازگاری روش با اکثر فضاهای رنگی<br>• بررسی ترکیب اولیه ویژگی های گابور و اطلاعات کانال های رنگی | • بار محاسباتی متوسط به دلیل محاسبه 12 ماتریس همرخدادی<br>• حساسیت به چرخش تصویر به دلیل عدم در نظر گرفتن تمامی جهات هشت گانه ماتریس همرخدادی<br>• عدم ارائه دلیل قطعی برای نادیده گرفتن اکثر ویژگی های آماری<br>• عدم ارائه دلیل قوی برای در نظر گرفتن ویژگی تجانس به عنوان |

[1] Homogeneity
[2] Support Vector Machine
[3] Red-Green-Blue
[4] Hue-Saturation-Value

| | |
|---|---|
| | ویژگی تفکیک کننده |

## 2-5-2- ترکیب رنگ و بافت بر پایه استخراج ویژگی های آماری از الگوهای دودویی محلی و ماتریس‌های هم‌رخدادی

اخلوفی و همکارانش در [101] روشی برای دسته بندی فرآورده های صنعتی از قبیل چوب، فابریک[1]، فیبرهای طبیعی [2] و تخته های سقفی [3] ارائه داده‌اند. با توجه به تصاویر رنگی اکثر فرآورده‌ها، ایشان در تحقیق خود تلاش کرده اند که از ویژگی های بافت در کنار رنگ، تواماً استفاده نمایند. همچون بخش قبل، در روش پیشنهادی اخلوفی و همکار ان نیز ابتدا تصویر رنگی (در هر فضای رنگی که باشد ) به سه کانال رنگی مجزا تقسیم می گردد. بنابراین روش ایشان هم در زمره‌ی گروه ترکیب اولیه قرار می‌گیرد. سپس در هر کانال رنگی به صورت مجزا ماتریس‌های همرخدادی (در هشت جهت متفاوت ) و الگوی دودویی محلی [4] محاسبه می‌گردد. در ادامه به ازای هر ماتریس همرخدادی، پنج ویژگی آماری شامل بی نظمی، انرژی، تباین، تجانس و همبستگی محاسبه می شود . از طرفی در تصویر خروجی ناشی از اعمال الگوی دودویی محلی، نیز همین ویژگی‌های آماری استخراج می گردد. نویسندگان برای مرحله دسته-بندی، از روش نزدیک ترین همسایه استفاده می کنند . ولیکن فاصله هر دو تصویر را طبق معادله زیر محاسبه می نمایند:

$$\Delta_{final} = \Delta_R + \Delta_G + \Delta_B \qquad (12-2)$$

در معادله فوق $\Delta_B$ , $\Delta_G$ , $\Delta_R$ به ترتیب نشان دهنده فاصله اقلیدسی بردارهای ویژگی استخراجی برای تصاویر در هر کدام از کانال های رنگی است. در مرحله نتایج، 4 پایگاه مجزا از تصاویر انواع چوب، فابریک، فیبر طبیعی و تخته سقفی به صورت دستی تهیه شده و در هر پایگاه عمل دسته بندی جداگانه انجام گرفته است . برای هر پایگاه دو آزمایش به صورت زیر انجام شده است :

- با استفاده از ویژگی‌های آماری استخراجی از ماتریس های همرخدادی در سه فضای رنگی مجزا RGB، HSL[5] و La*b*

[1] Fabric
[2] Organic Fibers
[3] Roofing Shingles
[4] Local Binary Patterns
[5] Hue-Saturation-Luminance

- با استفاده از ویژگی های آماری استخراجی از الگوهای دودویی محلی در سه فضای رنگی مجزا RGB، HSL و La*b*

در نهایت نتایج نشان داده است که در تمامی پایگاه ها بهترین نتایج در فضای رنگی RGB و با استفاده از ماتریس های همرخدادی حاصل شده است . به طور مثال در دسته بندی چوب ها، دقت 90٪ با استفاده از ماتریس همرخدادی و دقت 75٪ با استفاده از الگوی دودویی محلی ساده حاصل شد. به طور مشابه در مورد فیبرهای طبیعی نیز دقت دسته بندی بر پایه ماتریس همرخدادی 95٪ و بر پایه الگوی دودویی محلی 85٪ بدست آمد. مزایا و معایب روش [101] در جدول زیرمنعکس شده است.

**جدول 2-10. تحلیل مزیت ها و عیوب دسته بندی تصاویر رنگی بر اساس استخراج ویژگی های آماری از ماتریس های همرخدادی و الگوهای های دودویی محلی [101]**

| مزایا | معایب |
|---|---|
| • دقت دسته‌بندی نسبتاً بالا برای تصاویر فرآورده‌های صنعتی مورد آزمایش<br>• امکان سازگاری روش با اکثر فضاهای رنگی<br>• بررسی تواماً انواع ویژگی‌های آماری<br>• عدم حساسیت به چرخش به دلیل محاسبه ماتریس همرخدادی در هشت جهت و ذات چرخش ناپذیری الگوهای دودویی محلی | • بار محاسباتی بالا به دلیل محاسبه 24 ماتریس همرخدادی<br>• عدم تعمیم روش ارائه شده به تصاویر عمومی بافتی (نه فقط تصاویر برخی از فرآورده های صنعتی)<br>• عدم در نظر گرفتن نسخه های بهبود یافته الگوهای دودویی محلی (نسخه استفاده شده مربوط به 16 سال پیش است)<br>• نسخه الگوهای دودویی محلی در این مقاله، ذاتاً عملگری آماری نیست و واضح است که استخراج ویژگی‌های هارالیک از خروجی آن نتایج مثبتی نخواهد داشت |

## 2-5-3- ترکیب رنگ و بافت بر اساس نسخه بهبود یافته الگوهای دودویی محلی

پوربسکی و همکارانش در [102]، نسخه خلاقانه‌ای از الگوهای دودویی محلی را ارائه کرده اند که در آن ویژگی‌های بافت و رنگ به صورت تواماً در حین محاسبه‌ ی الگوهای دودویی محلی با یکدیگر ترکیب می‌شوند. بهتر آن است که قبل از بررسی این روش، به اختصار الگوهای دودویی محلی توضیح داده شود.

الگوی دودویی محلی [1] در واقع عملگری غیرپارامتریک است که تباین محلی و ساختار فضایی محلی تصویر را بررسی می‌کند. عملکرد این عملگر بدین صورت است که در ابتدا برای هر کدام

[1] Local Binary Patterns (LBP)

از پیکسل های تصویر، یک همسایگی در نظر گرفته می شود و سپس میزان الگوی دودویی محلی برای پیکسل مورد نظر در همسایگی تعیین شده ، به کمک معادله زیر محاسبه می شود

$$\boldsymbol{LBP_{P,R}} = \sum_{i=0}^{P-۱} S(g_i - g_c)۲^{i} \quad \text{به صورتی که} \rightarrow \quad S(x) = \begin{cases} ۱ & x \geq 0 \\ 0 & x < 0 \end{cases} \quad (13-2)$$

در معادله فوق $g_i$ شدت روشنایی نقاط همسایگی و $g_c$ شدت روشنایی نقطه مرکزی است . معمولاً برای آنکه عملگر فوق، نسبت به چرخش حساس نباشد، همسایگی به صورت دایره‌ای در نظر گرفته می‌شود. بنابراین با توجه به شعاع همسایگی (R)، تعداد نقاط همسایه (P) می تواند متفاوت باشد. چند نمونه از هم سایگی دایره ای با شعاع (R) در شکل(2-5) نشان داده شده است. همانطور که در شکل مشاهده می گردد، مختصات برخی از نقاط همسایگی دقیقاً روی مرکز پیکسل قرار نمی‌گیرند. شدت روشنایی این نقاط به کمک درون‌یابی محاسبه می شوند.

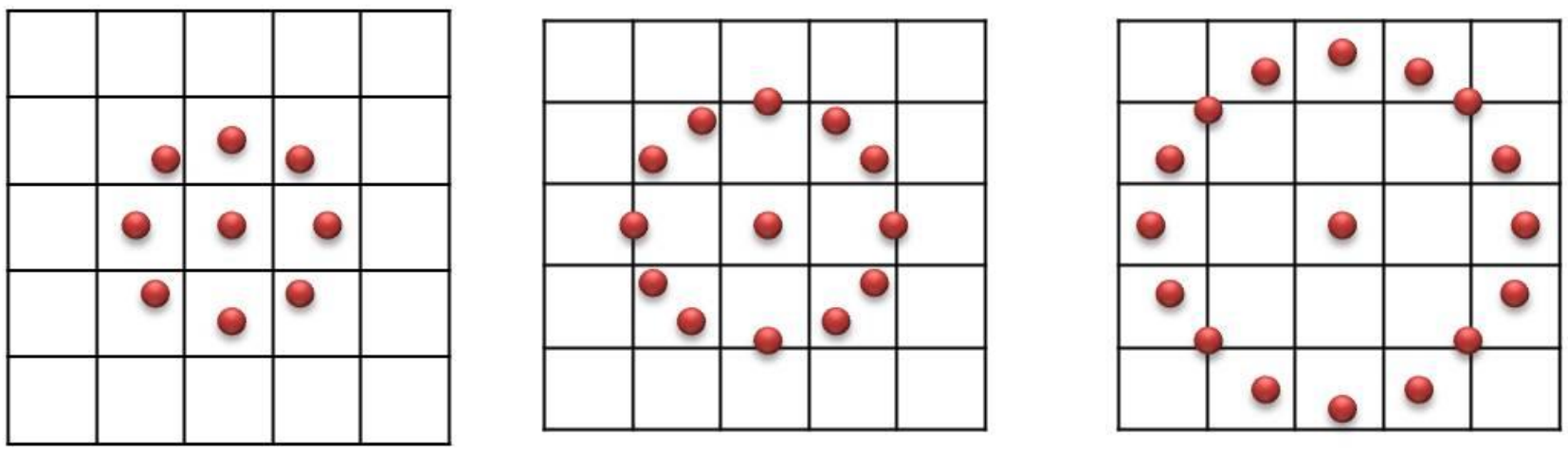

(الف) (ب) (ج)

شکل 2-5. نمونه هایی از همسایگی دایروی [108]

(الف). P= ۸ و R= ۱ (ب). P= 12 و R= 1.5 (ج). P= 16 و R=2

با توجه به تفاسیر فوق، خروجی عملگر الگوی دودویی محلی، عددی دودویی با P بیت اطلاعات است. در شکل زیر، مثالی از محاسبه الگوی دودویی محلی در نقطه مرکزی با همسایگی به شعاع 1، نشان داده شده است.

| | | |
|---|---|---|
| 49 | 217 | 203 |
| 172 | [redacted] | 166 |
| 94 | 85 | 47 |

( الف)

| | | |
|---|---|---|
| 0 | 1 | 1 |
| 1 | | 1 |
| 0 | 0 | 0 |

(ب)

| | | |
|---|---|---|
| 1 | 2 | 4 |
| 128 | | 8 |
| 64 | 32 | 16 |

(ج)

LBP=2+4+8+128=142

شکل2-6. مثالی از نحوه محاسبه الگوهای دودویی محلی

(الف) همسایگی مربعی 3×3 (ب) مقایسه پیکسل مرکزی با همسایه ها (ج) محاسبه LBP

در هر تصویر رنگی، هر پیکسل دارای سه مقدار شدت روشنایی مستقل در هر کانال رنگی به صورت مجزا است. پوربسکی و همکارانش در [102]، پیشنهاد دادند که درمعادله (2-13) هر سه مولفه رنگی همزمان در نظر گرفته شود . بنابراین معادله زیر جهت محاسبه میزان الگوی دودویی محلی را مطرح کردند :

$$\boldsymbol{CLBP_{P,R}} = \sum_{i=0}^{P-1} S\left(\sqrt{{C_{1n}}^2 + {C_{2n}}^2 + {C_{3n}}^2}\right) 2^i \quad \text{به صورتی که} \rightarrow \quad S(x) = \begin{cases} 1 & x \geq T \\ 0 & x < T \end{cases}$$

$$,T = \sqrt{{C_{1P}}^2 + {C_{2P}}^2 + {C_{3P}}^2} \qquad (14-2)$$

در معادله فوق، $C_{1P}$ و $C_{2P}$ و $C_{3P}$ به ترتیب نشان دهنده شدت روشنایی پیکسل مرکزی در کانال-های رنگی 1، 2 و 3 و $C_{1n}$ و $C_{2n}$ و $C_{3n}$ به ترتیب نشان دهنده شدت روشنایی همسایه در کانال‌های رنگی 1، 2 و 3 می باشد. نویسندگان در ادامه برای تصویر خروجی حاصل از اعمال عملگر الگوهای دودویی محلی، ویژگی های هارالیک را محاسبه و روش خود را بر روی پایگاه تصاویر رنگی BarkTex آزمایش نمودند. نتایج آنها، دقت 85.6 درصد را بر روی این پایگاه فراهم نمود. مزایا و معایب روش [102]، در جدول زیر بیان شده است.

**جدول 2-11. تحلیل مزیت ها و عیوب دسته بندی تصاویر رنگی بر اساس استخراج ویژگی های هارالیک از نسخه ترکیبی الگوهای دودویی محلی و اطلاعات کانال های رنگی[102]**

| مزایا | معایب |
|---|---|
| • دقت دسته‌بندی نسبتاً بالا بر روی پایگاه مورد نظر<br>• ترکیب اولیه اطلاعات رنگ و بافت که مزایای مطرح شده در مورد ترکیب اولیه را فراهم می‌کند | • بار محاسباتی نسبتاً بالا به دلیل محاسبه عملگر در سه کانال رنگی مجزا<br>• عدم ارائه دلیل قطعی برای موفقیت معادله ارائه شده بر روی تصاویر رنگی<br>• عملکرد عملگر ارائه شده می بایست بر روی دیگر پایگاه ها نیز بررسی شود |

## 2-5-4- ترکیب ثانویه رنگ و بافت بر پایه هیستوگرام الگوهای دودویی محلی و کانال‌های رنگی

برخلاف مثالهای پیشین، نینگ و همکارانش در [103] روشی ثانویه برای ترکیب ویژگی های رنگ و بافت ارائه دادند . در [103]، صورت مسئله اصلی نویسندگان ارائه روشی برای ردیابی اشیا در تصاویر ویدیویی است . بنابراین در مرحله یادگیری برای تعریف موقعیت و شکل شی مورد نظر تلاش کرده اند که ویژگی های رنگ و بافت را توأماً در نظر بگیرند تا ویژگی های

استخراجی معرف قوی‌تری از شی مورد نظر باشند. بنابراین سه هیستوگرام درجه‌بندی شده بر اساس ویژگی های رنگ در کانال های رنگی مجزا و یک هیستوگرام بر اساس نسخه اولیه الگوهای دودویی محلی تعریف کرده اند. در این روش، ابتدا تصویر رنگی (هر فضای رنگی دلخواه) به سه کانال مجزا تقسیم شده و سپس هیستوگرام نرمال شده هر کدام تشکیل می - گردد. سپس ستون های هر هیستوگرام را به دسته های n تایی تقسیم کرده و به اصطلاح هیستوگرام درجه‌بندی[1] می‌شود. در ادامه الگوی دودویی محلی به صورتی که پیش‌تر اشاره شد، محاسبه شده و هیستوگرام تصویر خروجی آن نیز نرمال و درجه بندی می‌شود. در نهایت هر دسته از هیستوگرام یکی از ابعاد بردار ویژگی نهایی را تشکیل داده و چهار هیستوگرام با یکدیگر پیوست[2] می‌شوند. بدین ترتیب اگر n=32 در نظر گرفته شود، یک بردار ویژگی 32 بعدی از تصویر ورودی استخراج خواهد شد . در بخش نتیجه گیری، نویسندگان بار محاسباتی پایین و دقت تشخیص بالا را از جمله مزایای روش خود ابراز می کنند. در این الگوریتم بر خلاف روش‌های پیشین، ویژگی‌های استخراجی رنگ و بافت به صورت مجزا استخراج شده و در نهایت بدون تلفیق، در کنار یکدیگر پیوست شدند. بنابراین این روش در زمره گروه ترکیب ثانویه قرار می گیرد. مطابق روال این مقاله مزایا و معایب این روش از دیدگاه نویسنده مقاله در جدول زیر نشان داده شده است.

**جدول 2-12. تحلیل مزیت ها و عیوب دسته بندی تصاویر رنگی بر اساس ترکیب ثانویه هیستوگرام کانال های رنگی و نسخه اولیه الگوهای دودویی محلی [103]**

| مزایا | معایب |
|---|---|
| • ترکیب ثانویه اطلاعات رنگ و بافت که در تصاویری که بافت و رنگ در سطح کلی تصویر ترکیب شده اند بسیار مفید است<br>• بار محاسباتی کمتر نسبت به برخی از روش های مبتنی بر هیستوگرام به دلیل استفاده از روش درجه بندی<br>• دقت نسبتا بالا در کاربرد مورد نظر | • حساسیت به نویزهای محلی به دلیل ضعف ماهوی نسخه اولیه الگوهای دودویی محلی و استخراج ویژگی‌های کلی رنگ<br>• عملکرد عملگر ارائه شده می بایست بر روی دیگر پایگاه ها نیز بررسی شود<br>• حساسیت به پارامترهای ورودی از جمله فضای رنگی و تعداد ستون ها در مرحله درجه بندی هیستوگرام |

[1] Quantized
[2] Concatenate

## 2-5-5- ترکیب رنگ و بافت بر پایه تابع تبدیل موجک[1] و شبکه عصبی[2]

همانطور که پیش تر نیز اشاره شد، استفاده از ویژگی های محیط فرکانس در حوزه آنالیز بافت تصاویر کاربرد وسیعی دارد. عبدالقدیر سنگور در [104]، روشی برای دسته‌بندی بافت تصاویر رنگی بر اساس ترکیب اولیه ویژگی های رنگ و بافت در محیط تبد یل موجک ارائه داده است. در این مقاله، ابتدا تصویر به سه کانال رنگی مجزا تقسیم شده و سه تصویر حاصل می شود. سپس برای هر کدام از تصاویر، توسط تابع تبدیل موجک سطح یکم [3]، سه زیر تصویر[4] (High-High) HH,(High-Low) HL,(Low-High) LH, تشکیل می گردد. در ادامه در هر زیر تصویر دو ویژگی بی نظمی و انرژی محاسبه می گردد. بنابراین در نهایت یک بردار ویژگی 18 بعدی تشکیل خواهد شد. در ادامه نیز از یک شبکه عصبی انتشار عقب گرد سه لایه استاندارد[5] جهت دسته بندی بافت تصاویر ورودی استفاده میگردد. در بخش نتایج [104]، یک پایگاه داده از 1920 تصویر رنگی از 16 کلاس بافتی تشکیل شده و دقت دسته‌بندی بر روی آن محاسبه گردیده است که به طور میانگین دقت 94.6 بدست آمده است. مزایا و معایب روش پیشنهادی در جدول زیر آورده شده است.

**جدول 2-13. تحلیل مزیت ها و عیوب دسته بندی تصاویر رنگی بر اساس ویژگی های آماری در زیر تصاویر موجک و شبکه عصبی انتشار عقبگرد[104]**

| مزایا | معایب |
|---|---|
| • دقت نسبتا بالا<br>• حساسیت اندک نسبت به نویز ضربه ای<br>• حساسیت اندک نسبت به تغییر مقیاس | • عدم کاربرد بالا در زمانی که اطلاعات رنگ و بافت در حد تصویر با یکدیگر ادغام شده اند<br>• بار محاسباتی نسبتا بالا به دلیل تغییر محیط |

## 2-5-6- روش ترکیبی پنج عملگر محلی و کلی برای ترکیب اطلاعات رنگ و بافت

خان و همکارانش در [14] یک روش ترکیبی فشرده برای استخراج همزمان اطلاعات بافت و رنگ ارائه داده اند . خان و همکارانش اعتقاد دارند که ترکیب ویژگی های استخراجی از عملگرهای انالیز رنگ و عملگرهای آنالیز بافت می تواند مفید باشد. به همین منظور در [14] پس از مطالعه دسته وسیعی از عملگرهای کلی و جزیی در حوزه آنالیز بافت و رنگ، پنج مورد

---

1 Wavelet Transform
2 Neural Network
3 First Order Wavelet
4 Sub-Band Image
5 Standard Three Layer Back-Propagation

را انتخاب کرده و بر روی تصویر اعمال کرده و پنج دسته بردار ویژگی مجزا استخراج می نمایند. سپس بردار های استخراجی را با روشی نوآورانه با یکدیگر ترکیب کرده و در نهایت با استفاده از روش های کاهش ابعاد [1] بهترین زیر مجموعه از ابعاد را جهت آنالیز تصاویر استفاده می‌نمایند. عملگرهای استفاده شده در این مقاله عبارتند از : الگوهای دودویی گابور [2]، رده بندی فاز محلی [3]، کد گشایی وبر [4]، الگوهای دودویی محلی کامل [5]، ویژگی های دودویی آماری [6]. در [14] دقت دسته بندی بر روی چهار پایگاه داده مشهور تصاویر بافتی رنگی به نام های FMD, KTH-TIPS-2a, KTH-TIPS-2b, Texture-10 آزمایش شده که نتایج بسیار خوبی فراهم کرده است.

**جدول 2-14. تحلیل مزیت ها و عیوب دسته بندی تصاویر رنگی بر اساس ترکیب ویژگی های استخراجی از پنج عملگر آنالیز محلی و کلی [14]**

| مزایا | معایب |
|---|---|
| • دقت دسته بندی بالا<br>• حساسیت اندک نسبت به چرخش به دلیل استفاده از برخی عملگرهای مستقل از چرخش | • بار محاسباتی بسیار بالا به دلیل اعمال همزمان پنج عملگر پیچیده<br>• حساسیت به نویز به دلیل عدم استقلال برخی از عملگرهای استفاده شده نسبت به انواع نویز<br>• عدم تضمین زیر مجموعه کاهش بعد یافته ویژگی‌ها برای دیگر پایگاه های داده |

## 2-5-7- ترکیب اطلاعات رنگ و بافت بر اساس ترکیب هیستوگرام رنگ و الگوهای دودویی محلی در محیط رنگی ترکیبی Ohta

پیتیکانن و همکارانش در [105] این ایده را مطرح می کنند که محیط های رنگی رایج به تنهایی نمی‌توانند معرف بافت های رنگی باشند، بنابراین اطلاعات رنگ تصویر در د و محیط رنگی RGB و Ohta را همزمان استخراج و ترکیب می نمایند. محیط رنگی Ohta که اولین بار توسط اوهتا در [106] مطرح شد، تبدیل خطی از محیط RGB است که شامل سه کانال رنگی مجزا به صورت زیر می باشد.

$$I1 = (R+G+B)/3, \quad I2=R-B, \text{ and } \quad I3 =(2G-R-B)2 \qquad (15 - 2)$$

در معادله فوق، R , G و B هر کدام نشان دهنده شدت روشنایی پیکسل مورد نظر در کانال های رنگی مجزا است . در [105] محققان دو راهکار متفاوت برای دسته بندی بافت تصاویر

[1] Feature Reduction
[2] Binary Gabor Pattern (BGP)
[3] Local Phase Quantization (LPQ)
[4] Weber Law encodes (WLE)
[5] Completed Local Binary Patterns (CLBP)
[6] Statistical Binary Features (SBF)

رنگی ارائه می دهند. درراهکار اول هیستوگرام تصویر در هر کانال رنگی از دو محیط فوق الذکر استخراج و رده بندی شده و در نهایت میزان رخداد هر رده [1] به عنوان یکی از ابعاد بردار ویژگی در نظر گرفته می شود . در راهکار دوم نیز الگوهای دودویی محلی بر روی هر کانال رنگی از دو محیط فوق به صورت مجزا پیاده سازی شده و در نهایت بردارهای ویژگی استخراجی با یکدیگر ترکیب می‌گردد. محققان در [105] روش ارائه شده خود را بر روی پایگاه داده‌های Outex و Vistex آزمایش کرده‌اند. نتایج تجربی پیتیکانن و همکارانش نشان می - دهد که استخراج هیستوگرام‌های کانال‌های رنگی در محیط Ohta بیشترین دقت دسته بندی را بر روی تصاویر پایگاه Outex فراهم می کند. همچنین استخراج هیستوگرام کانال های رنگی در محیط های RGB و Ohta به صورت برابر بیشترین دقت دسته بندی را بر روی پایگاه Vistex فراهم می‌کنند.

**جدول 2-15. تحلیل مزیت ها و عیوب دسته بندی تصاویر رنگی بر اساس ترکیب هیستوگرام رنگ و الگوهای دودویی محلی در محیط رنگی ترکیبی Ohta [105]**

| مزایا | معایب |
|---|---|
| • دقت دسته بندی مناسب<br>• معرفی توانایی انواع محیط های رنگی در معرفی بافت تصاویر<br>• بار محاسباتی نه چندان بالا | • حساسیت به نویز به دلیل استفاده از هیستوگرام<br>• عدم درنظر گرفتن ویژگی های محلی<br>• عدم تضمین روش ارائه شده برای انواع پایگاه های داده |

## 2-5-8- ترکیب اطلاعات رنگ و بافت بر اساس تبدیل ویژگی مستقل از مقیاس و اطلاعات کانال های رنگی

همانطور که در زیر بخش 2-1-2 اشاره شد، یکی از عملگرهای موفق در آنالیز بافت تصویر، عملگر SIFT است. نسخه ابتدایی SIFT برای آنالیز تصاویر با سطوح خاکستری مطرح شد ولیکن در [140]، عبدالحکیم و فاراگ ، با ترکیب ویژگی های استخراجی از SIFT و اطلاعات کانال‌های رنگی ، نسخه بهبود یافته آن را تحت عنوان CSIFT مطرح کردند. در [140]، نویسندگان محیط رنگی جدیدی تحت عنوان مدل رنگی گوسین تعریف می نمایند که به صورت تبدیل خطی‌ای از تصویر در فضای رنگی RGB قابل نمایش است. نحوه جابجایی تصویر به فضای رنگی جدید در معادله زیر نشان داده شده است.

$$\begin{Bmatrix} E \\ E_{\lambda} \\ E_{\lambda\lambda} \end{Bmatrix} = \begin{Bmatrix} 0.06 & 0.63 & 0.27 \\ 0.3 & 0.04 & -0.35 \\ 0.34 & -0.6 & 0.17 \end{Bmatrix} \begin{Bmatrix} R \\ G \\ B \end{Bmatrix} \quad (16-2)$$

[1] Quantized Bin

در نهایت نیز، تصویر در محیط جدید با فیلتر گوسین با پارامترهای ورودی قابل تنظیم، فیلتر می‌شود. نویسندگان در [140]، ادعا می کنند که تصویر فیلتر شده در این محیط نسبت به اطلاعات کانال های رنگ مستقل شده است . در [140]، از CSIFT برای حل مسئله تشخیص شی در قالب یک الگوریتم سه مرحله ای استفاده شده است که به شرح زیر است:

- تشخیص نقاط ورودی[1]
- تولید عملگر (توضیح دهنده )[2]
- تنظیم ویژگی و تخمین ژست[3] (جهت)

در [140] به صورت مستقیم از CSIFT برای حل مسئله دسته بندی بافت تصاویر رنگی استفاده نشده است ولیکن توانایی بالای آن در استخراج ویژگی های بافتی و آنالیز بافت تصویر، باعث شده که مورد توجه محققان این حوزه قرار گیرد . در این فصل برخی از کاربردی ترین روش‌های آنالیز بافت و رنگ مطرح و مورد بررسی قرار گرفت . در برخی تحقیقات، نویسندگان تلاش کرده‌اند که از اطلاعات ترکیبی بافت و رنگ برای کاربردهای گوناگون پردازش تصویر همچون آنالیز تصاویر پزشکی شامل غدد، عروق خونی، بازرسی بصری سطوح شامل چوب، پارچه و سنگ، بازیابی تصاویر، تشخیص دود و آتش، تشخیص حرکت، ردیابی اشیا استفاده کنند. به همین دلیل تحقیقات متعدد و متنوعی برای آنالیز و دسته بندی بافت تصاویر رنگی ارائه شده است که توضیح هر کدام مفصل بوده و در حوصله این مقاله نمی گنجد. ولیکن به برخی از کارآمدترین این روش‌ها در جدول زیر اشاره شده است.

[1] Inserting Points Detection
[2] Descriptor Building
[3] Feature Matching and Pose Estimation

**جدول 2-16. بررسی اجمالی برخی از روش های ارائه شده جهت آنالیز و دسته بندی بافت تصاویر رنگی**

| ردیف | سال ارائه | روش ارائه شده | مرجع |
|---|---|---|---|
| 1 | 2002 | ترکیب هیستوگرام رنگ و الگوهای دودویی محلی در محیط رنگی ترکیبی Ohta | [105] |
| 2 | 2005 | مدل پنهان مارکوف[1] در فضای موجک | [120] |
| 3 | 2006 | ترکیب اطلاعات عملگر تبدیل ویژگی مستقل ازمقیاس و اطلاعات کانال های رنگی (CSIFT) | [140] |
| 4 | 2007 | ترکیب اطلاعات عملگر تبدیل ویژگی مستقل ازمقیاس با هیستوگرام های ماتریس های همرخدادی رنگی (SIFT-CCH) | [141] |
| 5 | 2008 | ویژگی های آماری از الگوهای دودویی محلی و ماتریس های همرخدادی | [101] |
| 6 | 2008 | ویژگی های هارالیک از نسخه ترکیبی الگوهای دودویی محلی و اطلاعات کانال های رنگی | [102] |
| 7 | 2009 | ترکیب هیستوگرام کانال های رنگی و نسخه اولیه الگوهای دودویی محلی | [103] |
| 8 | 2009 | ویژگی های آماری در زیر تصاویر موجک و شبکه عصبی انتشار عقبگرد | [104] |
| 9 | 2011 | الگوهای دودویی محلی اولیه در محیط ترکیبی خطی ORGB | [122] |
| 10 | 2011 | ترکیب ویژگی های استخراجی از الگوهای دودویی محلی در محیط های رنگی HSV , YCbCr , oRGB | [122] |
| 11 | 2011 | ترکیب ویژگی های استخراجی از الگوهای دودویی محلی اولیه در تصویر با سطوح خاکستری و تصاویر رنگی در کانال های مختلف رنگ | [122] |
| 12 | 2012 | تفاصل نیمه کامل مبتنی بر محاسبات[2](PDE) برای تجزیه مدل تجربی[3] (EMD) | [135] |
| 13 | 2012 | الگوهای برداری رنگ محلی چند مقیاسی | [130] |
| 14 | 2012 | ترکیب عملگر تبدیل ویژگی‌های مستقل از مقیاس[4] (SIFT) استخراجی از پنجر-هایی با سایز مختلف و عملگر آماری اسپکتروم چند فرکتالی[5] | [131] |
| 15 | 2013 | انتخاب زیر مجموعه ای از ویژگی‌های استخراجی از فضاهای رنگی مختلف | [117] |
| 16 | 2014 | ماتریس های همرخدادی در سطح رنگ | [10] |
| 17 | 2014 | ترکیب پنج عملگر آنالیز محلی و کلی (الگوهای دودویی گابور، رده بندی فاز محلی، کد گشایی وبر، الگوهای دودویی محلی کامل، و ویژگی‌های دودویی آماری) | [14] |
| 18 | 2015 | ترکیب مدل پسرفت خودکار[6] و تجزیه مدل تجربی دو بعدی[7] (BEMD) | [132] |
| 19 | 2016 | الگوی مقاوم به نویز کامل با ساختار محلی[8] و با ساختار کلی | [133] |
| 20 | 2016 | نسخه بهبود یافته عملگر محلی وبر[9] (WLD) | [134] |

[1] Hidden Markov Model
[2] Partial Differential Equation-based
[3] Emprical Mode Decomposition
[4] Scale-Invariant Feature Transform (SIFT)
[5] Multi Fractal Spectrum
[6] Autoregressive Model
[7] Bidimensional Emprical Mode Decomposition (BEMD)
[8] Completed Noise-invariant Local-structure Patterns (CNLP)
[9] Weber Local Descriptor (WLD)

## 2.6. Local binary patterns and it's improved versions

همانطورکه پیش‌تر در بخش 2-5-3 مطرح شد، الگوهای دودویی محلی یکی از عملگرهای بسیار قوی برای آنالیز بافت تصویر می باشد. در این مقاله در جهت استخراج ویژگی های بافتی تصاویر، از این عملگر الهام گرفته شده اس ت. همانطور که در فصل آتی به تفصیل شرح داده خواهد شد، نسخه بهبود یافته ای از این عملگر با تمرکز بر حل محدودیت های آن از جمله حساسیت نسبت به نویز ضربه‌ای و عدم تفکیک الگوهای معنی‌دار و بی معنی ارائه خواهد شد. بنابراین در این زیر بخش نسخه ابتدایی این عملگر و برخی از نسخه های بهبود یافته آن مرور خواهد شد.

### 2.6.1. Basic Version of Local Binary Patterns

نسخه اولیه الگوهای دودویی محلی برای اولین بار توسط اوجالا و همکارانش در [107]، جهت آنالیز و طبقه‌بندی بافت تصویر ارائه شد. الگوهای دودویی محلی در واقع مجموعه عملگرهایی غیرپارامتریک هستند که تباین محلی [1] و ساختار فضایی محلی [2] تصویر را آنالیز می نمایند. عملکرد این عملگر بدین صورت است که در ابتدا برای هر کدام از پیکسل های تصویر، یک همسایگی به مرکزیت پیکسل و شعاع های متفاوت، در نظر گرفته شده و سپس مقدار شدت روشنایی هر کدام از همسایگان با پیکسل مرکزی مقایسه می گردد و در نهایت میزان الگوی دودویی محلی پیکسل مورد نظر در قالب یک الگوی دودویی به صورتی که در معادله زیر نشان داده شده است، سنجیده می شود.

$$\boldsymbol{LBP_{P,R}} = \sum_{k=0}^{P-1} S(g_k - g_c)2^k \text{ به صورتی که } \rightarrow \quad S(x) = \begin{cases} 1 & x \geq 0 \\ 0 & x < 0 \end{cases} \quad (17-2)$$

در معادله (2-17) به ترتیب، P تعداد نقاط همسایه و $g_k$ شدت روشنایی نقاط همسایگی و $g_c$ شدت روشنایی نقطه مرکزی است. به طور معمول برای آنکه عملگر فوق، نسبت به چرخش حساس نباشد، همسایگی به صورت دایره ای در نظر گرفته می شود. چند نمونه از همسایگی دایروی با شعاع (R) در شکل (2-5) نشان داده شده است. خروجی عملگر الگوی دودویی محلی، عددی دودویی با P بیت اطلاعات است. همانطور که در شکل (2-3)، نشان داده شده، نحوه اندیس‌گذاری پیکسل‌های همسایه به شدت بر روی عدد مبنای ده الگوی دودویی بدست آمده، تاثیر گذار است و می تواند منجر به تغییر مقدار الگوی دودویی محلی گردد. همانطور که در شکل (2-7) مشاهده میشود، پس از مقایسه شدت روشنایی هر پیکسل همسایه با پیکسل

---
[1] Local Contrast
[2] Local Spatial Structure

مرکزی، الگویی دودویی به صورتی که در شکل (ب) نشان داده شده است، استخراج می گردد. برای تبدیل الگوی دودویی استخراجی به عددی در مبنای ده، می بایست از بخش اول فرمول 2-17 استفاده گردد. ولیکن در این فرمول محل آغاز اندیس گذاری مشخص نشده است، بنابراین به هشت طریق مختلف می توان این عمل را انجام داد که دو نمونه آن در شکل های (ج) و (د) نشان داده شده است. در نهایت همانطور که مشاهده می شود، عدد محاسبه شده در مبنای ده به شدت به نحوه اندیس‌گذاری وابسته است.

| | | |
|---|---|---|
| 49 | 217 | 203 |
| 172 | | 166 |
| 94 | 85 | 47 |

(الف)

| | | |
|---|---|---|
| 0 | 1 | 1 |
| 1 | | 1 |
| 0 | 0 | 0 |

(ب)

| | | |
|---|---|---|
| 1 | 2 | 4 |
| 128 | | 8 |
| 64 | 32 | 16 |

(ج)

| | | |
|---|---|---|
| 4 | 8 | 16 |
| 2 | | 32 |
| 1 | 128 | 64 |

(د)

LBP (ج) = 2+4+8+128= 142

LBP (د) = 2+8+16+32= 58

شکل2-7. نحوه اندیس گذاری های گوناگون در الگوی دودویی محلی

(الف) همسایگی مربعی 3×3 (ب) مقایسه پیکسل مرکزی با همسایه ها (ج)یکی از نحوه های اندیس گذاری و محاسبهLBP (د) یکی از نحوه های اندیس گذاری و محاسبهLBP

برای حل این معضل، اوجالا و همکارانش در [66] پیشنهاد دادند که با چرخش عدد دودویی بدست آمده و انتخاب کمینه مقادیر مم کن، می توان مقدار یکتایی را به هر کدام از الگوهای محلی اختصاص داد. این موضوع در معادله (2 - 18) نشان داده شده است.

$$\boldsymbol{LBP_{P,R}^{ri}} = \min\{ROR\{LBP_{P,R}, \alpha)\ |\ \ \alpha = 0,1,\ldots,p-1\ \} \quad (18-2)$$

در معادله (2-18)، عدم حساسیت اپراتور نسبت به چرخش با نماد "ri" نشان داده شده است. همچنین "ROR" نشان دهنده چرخش به سمت راست[1] است که "α"بار تکرار شده و حداقل اعداد بدست آمده به ازای "α"های بین صفر تا p-1 به عنوان الگوی دودویی محلی در نظر گرفته می شود. یکی از سوالات اصلی که پس از مطالعه نسخه اولیه الگوهای دودویی محلی، به ذهن متصور می‌شود، چگونگی آنالیز تصویر توسط الگوی دودویی محلی است . به عبارت دیگر چرا الگوهای دودویی محلی می توانند معرف خوبی از بافت تصویر باشند و جهت آنالیز بافت استفاده شوند.

همانطور که در معادله (2 - 17) دیده شد، همسایگانی که شدت روشنایی آنها بزرگتر و یا مساوی با پیکسل مرکزی باشند، برچسب "یک" و همسایگانی که شدت روشنایی ایشان کمتر

---

[1] Rotate Right

از شدت روشنایی مرکز باشند، برچسب "صفر" خواهند گرفت. بنابراین برای هر همسایگی با شعاع R و تعداد P همسایه، حالات بسیار گوناگونی از الگوهای دودویی امکان رخ دادن دارد. به طور مثال استفاده از عملگر الگوی دودویی با شعاع 1 و 8 همسایه، می تواند منجر به 36 حالت گوناگون از الگوهای دودویی شود . تمامی این حالات در شکل زیر نشان داده شده است . این شکل از منبع [66] جهت بررسی نحوه کارایی الگوهای دودویی محلی استخراج و آورده شده است.

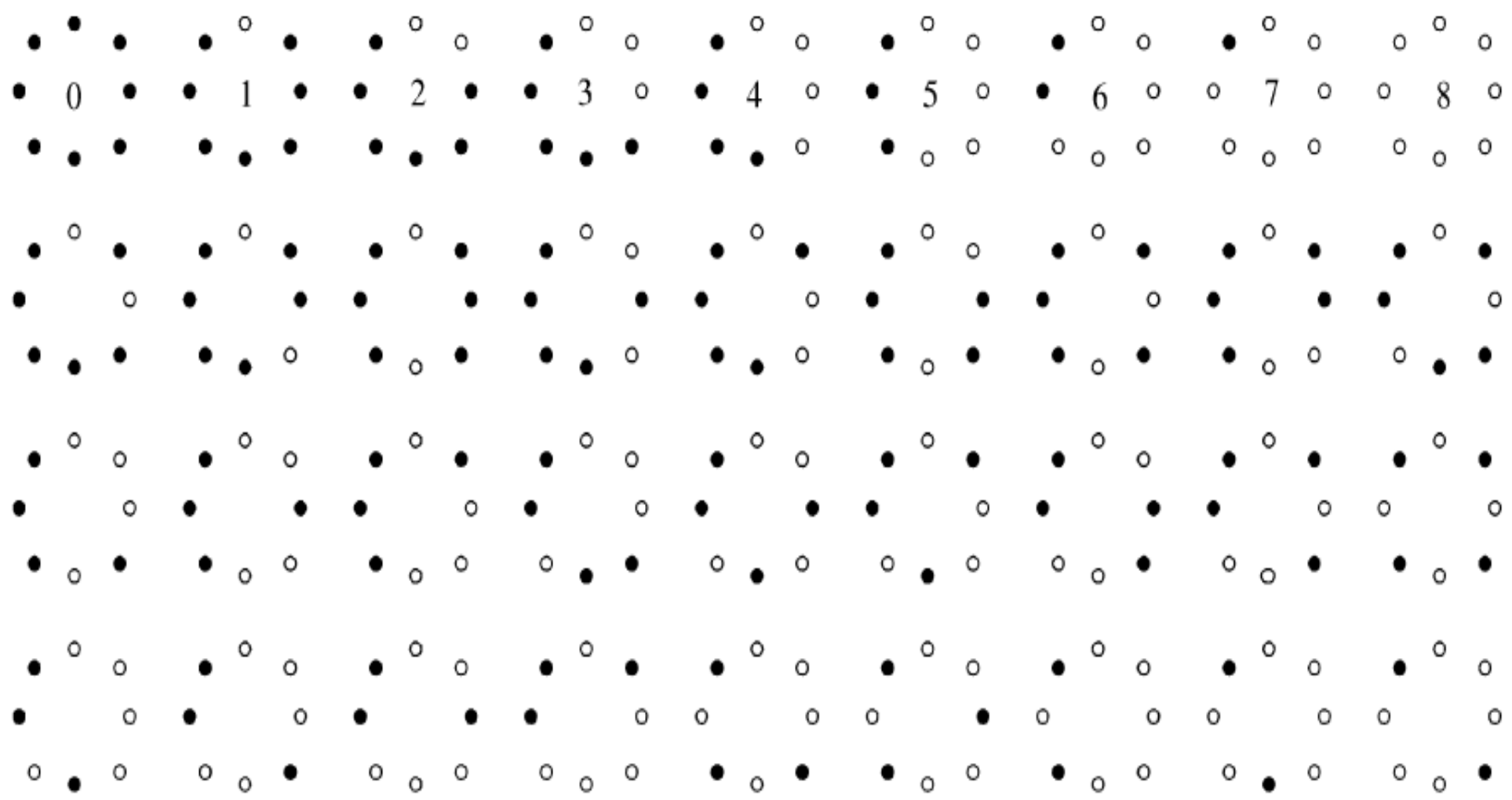

شکل 2-8. 36 الگوی دودویی ممکن با استفاده از $LBP_{8,1}$. نقاط مشکی نشاندهنده صفر و نقاط سفید معرف یک هستند [66]

با توجه به شکل (2-8) می‌توان به این نتیجه رسید که هر کدام از حالات فوق معرف ویژگی‌های خاصی از تصویر و نحوه ارتباط بین پیسکل‌ها با همسایگانشان هستند. به طور مثال الگوی نشان داده شده با شماره 4 معرف حالت لبه در تصویر است. در مثالی دیگر الگوی هشت معرف وجود خال‌های تاریک و دشت‌ها (بخشی محلی با شدت روشنایی تقریبا ثابت ) در تصویر است. الگوی شماره صفر نیز قدرت تشخیص خال های روشن را دارد . فرم های شماره سه و پنج نیز نقاط گوشه را آشکار می کنند. فرم شماره دو و شش نقاط منتهای خطی را تشخیص می دهند. الگوی شماره یک نیز خال ضعیف را تشخیص می دهد. با این تفاسیر، اکثر حالاتی که توسط عملگر الگوی دودویی محلی تشخیص داده شده و از حالات دیگر تمیز داده می شود، معرف ارتباط خاصی مابین پیکسل مرکز و تصویر است . بنابراین مشخص است که عملگر الگوی دودویی محلی، قدرت تشخیص انواع الگوهای محلی در تصویر را دارد و در نهایت با دسته بندی آنها عملاً بافت تصویر را آنالیز می کند.

## 2-6-2- الگوهای دودویی محلی بهبود یافته[1]

نتایج آزمایشات پیتیکئینن[2] و همکارانش در [65]، نشان داد که الگوی دودویی محلی اولیه به صورتی که در بخش قبل توضیح داده شد، علي الرغم توانایی مناسب برای آنالیز بافت تصویر، بار محاسباتی بالایی داشته و قدرت جداسازی [3] مناسبی را برای انواع تصاویر بافتی فراهم نمی آورد. بنابراین پس از مدتی، شکل بهبود یافته ی عملگر الگوی دودویی محلی توسط اوجالا [4] و پیتیکینن در [66] ارائه شد. در الگوی دودویی محلی بهبود یافته، معیاری به نام میزان یکنواختی[5] (همگنی) طبق معادله (2 - 19) تعریف می شود.

$$U(LBP_{P,R}) = \left|s\left(g_{p-1} - g_c\right) - s(g_0 - g_c)\right| + \sum_{k=1}^{p-1}\left|s(g_k - g_c) - s(g_{k-1} - g_c)\right| \quad (19-2)$$

همانگونه که در معادله فوق دیده می شود، میزان یکنواختی نشان دهنده تعداد جهش ها (جابجایی از صفر به یک و بالعکس) در شدت روشنایی نقاط همسایگی پشت سر هم است . به طور مثال در الگوی "00011011" میزان یکنواختی برابر با 4 و در "00000001" میزان یکنواختی برابر با 2 است. سپس حد آستانه یکنواختی ( $U_T$) توسط کاربر تعیین می گردد و الگوهایی که میزان یکنواختی آنها کمتر از حدآستانه یکنواختی باشد، الگوهای یکنواخت و الگوهایی که میزان یکنواختی آنها بیش از $U_T$ باشد، به عنوان الگوهای غیر یکنواخت تعریف می شوند. درنهایت نیز با توجه به این تعریف، میزان الگوی دودویی محلی بهبود یافته طبق معادله(2 - 20) محاسبه می شود.

$$MLBP_{P,R}^{riu_T} = \begin{cases} \sum_{k=0}^{P-1} s(g_k - g_c) & if\ U(LBP_{P,R}) \le U_T \\ P+1 & \text{در غیر این صورت} \end{cases} \quad (20-2)$$

همانگونه که از معادله فوق برمی‌آید، در شکل بهبود یافته‌ی الگوی دودویی محلی، به همسایگی های یکنواخت، برچسب هایی[6] از صفر تا P و به همسایگی های غیریکنواخت برچسب 1+P اختصاص داده می شود. با توجه به اینکه در الگوی دودویی محلی بهبود یافته به تمام همسایگی‌های غیریکنواخت برچسب یکسان الصاق می‌شود، بنابراین برچسب‌های الصاق شده به الگوهای یکنواخت موجود در تصویر باید اکثر الگوهای موجود در تصویر را پوشش دهد و الگوهای غیر یکنواخت تنها بخش ناچیزی از الگوها را شامل شود . نتایج عملی تاجری پور و همکارانش در [15، 16 و 108] نشان داد که در اکثر کاربردهای پردازش تصویر (مانند

---

[1] Modified Local Binary Patterns (MLBP)
[2] Pietikeinnen
[3] Discrimination
[4] Ojala
[5] Uniformity Measure
[6] Labels

سیستم‌های بازرسی بصری، بازیابی تصویر، تشخیص عیوب سطحی و ...) چنانچه حد آستانه میزان یکنواختی ( $U_T$) برابر با P/4 در نظر گرفته شود، درصد کمی از الگوها (کمتر از یک درصد) دارای برچسب غیریکنواخت خواهند بود و قدرت تشخیص بالاتر برود . در این مقاله همچون برخی از مقالات، این نسخه از الگوهای دودویی محلی که از معیار یکنواختی برای برچسب گذاری پیکسل‌های تصویر استفاده می‌کند، را با نماد $MLBP_{P,R}$ نشان خواهیم داد.
با توجه به محدودیت ها و نقاط ضعفی که اشاره شد، تاکنون نسخه های بهبود یافته گوناگونی از الگوهای دودویی محلی در حوزه های مختلف مطرح شده است که غالبا تلاش کرده اند یکی از محدودیت‌ها را حل نمایند یا کیفیت این عملگر را در حوزه های کاربردی مختلف افزایش دهند. ذکر تمامی این روش ها همراه با جزییات ، امری به شدت زمان بر می‌باشد که گاها در حوزه تخصصی این مقاله (دسته بندی تصاویر بافتی رنگی) نیز نمی‌گنجد. ولیکن ما در همین راستا تاکنون تحقیقات گوناگونی را انجام داده‌ایم که در زیر به برخی از آنها اشاره خواهد شد .
شایان ذکر است که بار محاسباتی MLBP حتی در همین شرایط نسبت به بسیاری از عملگرهای آنالیز بافت تصویر کمتر است.

## 2.6.3. One dimensional local binary pattern

همانگونه که از توضیحات فوق بر می آید، همسایگی در عملگر بهبود یافته ی الگوی دودویی محلی به صورت دو بعدی انتخاب می گردد، بنابراین این نسخه از الگوهای دودویی محلی را الگوی دودویی محلی دو بعدی نیز می نامند. با توجه به همین موضوع (انتخاب همسایگی دو بعدی)، بار محاسباتی MLBP از درجه دوم است که پیچیدگی محاسباتی سیستم را افزایش می‌دهد. در الگوی دودویی محلی دو بعدی، انتخاب همسایگی به صورت دایروی، باعث عدم حساسیت اپراتور نسبت به چرخش تصویر می شود. ولیکن در بر خی از کاربرد های پردازش تصویر همچون سیستم های بازرسی بصری [16]، چرخش تصویر اهمیت بالایی ندارد، بنابراین نیازی به انتخاب همسایگی دایروی نیست. از طرفی محاسبات درون یابی مورد نیاز در همسایگی‌های دایره‌ای بار محاسباتی بالایی را به سیستم وارد می کند که این امر می‌تواند توان برخط[1] بودن سیستم را به شدت پایین بیاورد در حالی که با توجه به توضیحات پیشین، یکی از اهداف این مقاله طراحی سیستمی برخط جهت کاربرد عینی در صنایع است . بنابراین در این بخش نسخه جدیدی از الگوی دودویی محلی معرفی شده است که در آن همسایگی به صورت قطعه[2] افقی(عمودی) در نظر گرفته می شود. با توجه به انتخاب همسایگی به صورت قطعات افقی (عمودی)، این نسخه الگوی دودویی محلی عملگری از درجه اول خواهد بود . بنابراین از

---

[1] Online
[2] Row(Column) Segment

این به بعد، این نسخه از الگو های دودویی محلی را الگوی یک بعدی می نامیم . در الگوی دودویی محلی یک بع‌دی، سطح خاکستری اولین پیکسل از تصویر به ترتیب با سطوح خاکستری دیگر پیکسل‌های موجود در آن قطعه مقایسه می شود. مثالی از اعمال عملگر الگوی دودویی محلی یک بعدی بر روی یک تکه عمودی از تصویر در شکل زیر نشان داده شده است

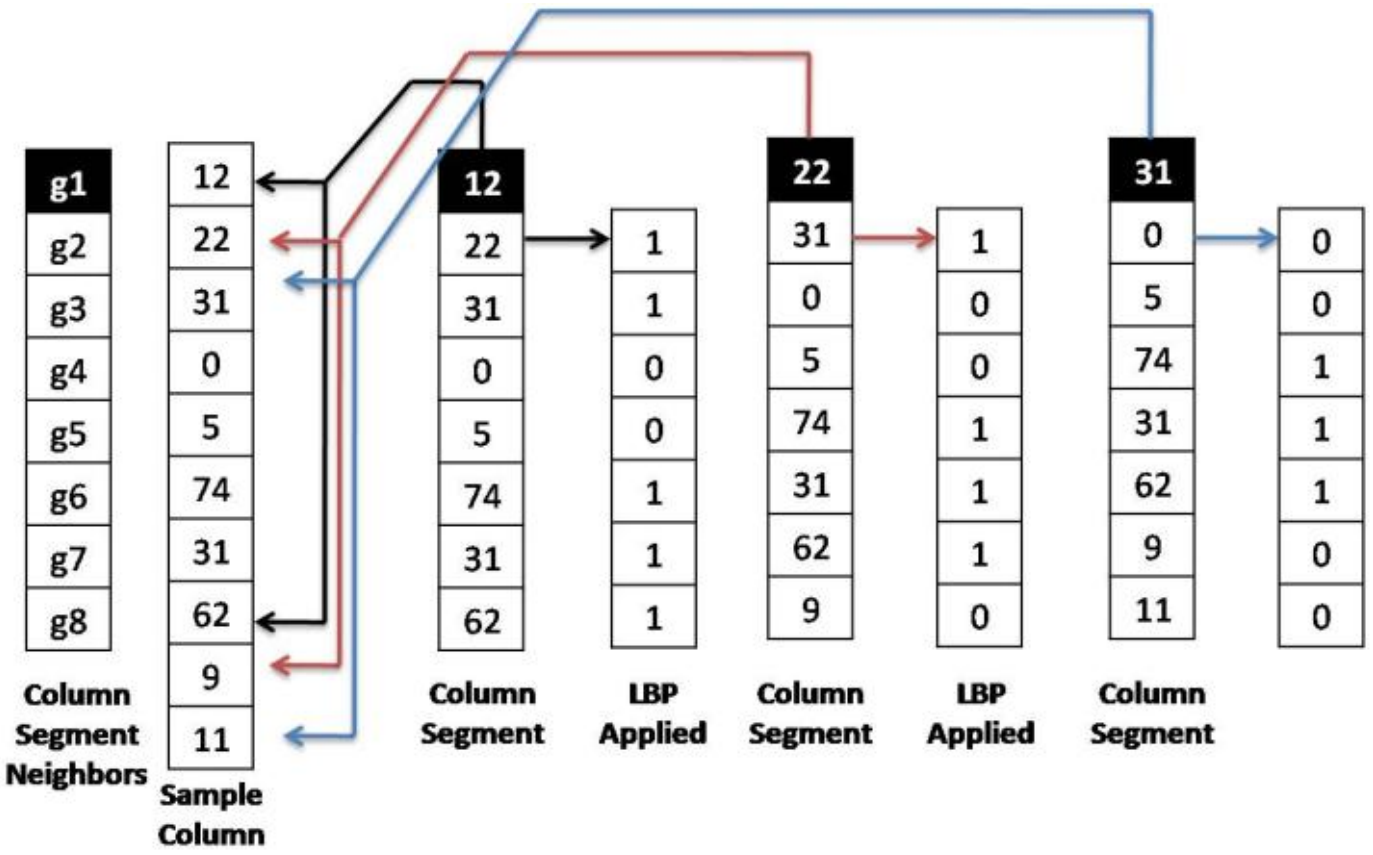

شکل2-9. اعمال عملگر الگوی دودویی محلی بر روی تکه ای عمودی از تصویر با قطعه ای عمودی به طول 8 [16]

در این روش میزان یکنواختی برابر با تعداد جهش ها (جابجایی از صفر به یک و بالعکس) در شدت روشنایی نقاط همسایه در قطعه افقی(عمودی) تعریف می شود. معادله (2 - 21)، این مطلب را نشان می دهد.

$$\boldsymbol{U(LBP_L)} = \left|s\left(g_L - g_1\right) - s(g_2 - g_1)\right| + \sum_{i=2}^{L-1}\left|s\left(g_i - g_1\right) - s(g_{i+1} - g_1)\right| \quad (21-2)$$

در معادله فوق، $L$ طول قطعه افقی (عمودی) را بر حسب پیکسل نشان می دهد و $g_1$ نشان دهنده سطح خاکستری اولین پیکسل در آن قطعه است . همچنین به دلیل تغییر شکل همسایگی، نماد $LBP_{P,R}$ به فرم $LBP_L$ تغییر می یابد. شایان ذکر است که اگر طول $L$ بیش از اندازه کوچک در نظر گرفته شود، الگوهای استخراجی نمی توانند معرف رخدادهای خاص و ویژگی‌های خاص محلی در تصویر باشند و اگر طول $L$ بیش از اندازه بزرگ در نظر گرفته شود، علاوه بر افزایش بار محاسباتی، ارتباط معنی داری مابین پیکسل های همسایه بسیار دور و پیکسل آغازین وجود نخواهد داشت و ویژگی های محلی تصویر از دست می رود. در این نسخه نیز الگوهایی که میزان یکنواختی آنها کمتر از حدآستانه ی میزان یکنواختی ($U_T$ ) باشد، الگوهای یکنواخت و الگوهایی که میزان یکنواختی آنها بیش از $U_T$ باشد، به عنوان الگوی غیر یکنواخت تعریف می شوند . بنابراین با توجه به این تعریف، میزان الگوی دودویی محلی یک بعدی برای قطعات افقی (عمودی) طبق معادله زیر محاسبه می شود.

$$LBP_L^{U_T} = \begin{cases} \sum_{i=۲}^{L} S(g_i - g_۱) & if\ U(LBP_L) \leq U_T \\ L & \text{شرایط دیگر} \end{cases} \qquad (22-2)$$

همانگونه که از معادله فوق، برمی آید، در الگوی دودویی محلی یک بعدی، به قطعات افقی(عمودی) یکنواخت، برچسب هایی از صفر تا L-1 و به قطعات افقی (عمودی) غیریکنواخت برچسب L اختصاص داده می شود. برچسب های الصاق شده به الگوهای یکنواخت موجود در تصویر باید اکثر الگوهای موجود در تصویر را پوشش دهد و الگوهای غیر یکنواخت تنها بخش ناچیزی از الگوها را شامل شود . الگوهای دودویی محلی دو بعدی به دلیل در نظر گرفتن همسایگی دو بعدی، عملگری از درجه دوم است ولیکن عملگر الگوی دودویی محلی یک بعدی به دلیل در نظر گرفتن همسایگی به صورت قطعات افقی و عمودی، عملگری از درجه اول است. به همین دلیل پیچیدگی محاسباتی آن نسبت به الگوهای دودویی محلی دو بعدی بسیار کمتر است. اثبات این مسئله در بخش بعدی به تفصیل بررسی شده است.

## 2-6-4 - بازیابی تصویر[1] به کمک الگوهای دودویی محلی بهبود یافته

همانطور که توضیح داده شد، محققین تلاش کرده اند که از الگوهای دودویی محلی در انواع مسائل حوزه پردازش تصویر استفاده کنند تا بیشترین دقت بدست آید . محققین در [15] از الگوهای دودویی محلی برای بازیابی تصویر استفاده کرده است . بازیابی تصویر به معنای استخراج نزدیک‌ترین نمونه از درون پایگاه داده بزرگی از تصاویر به تصویر پرس وجو[2] است که توسط کاربر ارائه شده است. همانطور که از این تعریف بر می‌آید، اصلی‌ترین مرحله در مبحث بازیابی تصویر، ارائه روشی برای سنجش میزان شباهت تصاویر به یکدیگر است . همانطور که بررسی شد الگوهای دودویی محلی بهبود یافته یکی از روش‌های معتبر در حوزه آنالیز بافت تصویر است. جهت شناساندن بافت تصویر به سیستم های کامپیوتری و مقایسه تصاویر با یکدیگر، نتایج آنالیز بافت تصویر توسط عملگر فوق می بایست به صورت برداری عددی تبدیل گردد. در [15] روشی مطمئن برای تبدیل نتایج آنالیزی الگوهای دودویی محلی به بردار ویژگی عددی ارائه شده است. پس از اعمال عملگر $MLBP_{P,R}$ بر روی تصویر، به هر همسایگی محلی یکنواخت برچسب هایی از صفر تا P و به الگوهای غیریکنواخت برچسب P+1 اختصاص داده می شود . بنابراین در مرحله استخراج بردار ویژگی می توان برای هر تصویر ورودی یک

---

[1] Image Retrieval
[2] Query

بردار P+2 بعدی استخراج کرد. برای استخراج بردار ویژگی، پس از اعمال عملگر بر روی تصویر و اختصاص برچسب ها به همسایگی ها، احتمال رخداد هر کدام از برچسب ها به عنوان یکی از ابعاد بردار ویژگی محاسبه می گردد. احتمال رخداد هر برچسب به صورت نسبت تعداد پیکسل-ها (همسایگی‌ها) با آن برچسب به تعداد کل پیکسل ها (همسایگی‌ها) تعریف می شود. این مطلب در معادله زیر نشان داده شده است.

$$\boldsymbol{f_i} = \frac{N_{f_i}}{N_{total}} \qquad 0 \le i \le P+1 \qquad (23-2)$$

در معادله فوق، $f_i$ احتمال برخورد به برچسب "i" و $N_{fi}$ تعداد کل پیکسل‌ها با برچسب "i" است. $N_{total}$، نیز تعداد کل پیکسل های تصویر است که با ضرب اندازه سطر ها در ستون های تصویر بدست می آید. در معادله زیر بردار ویژگی نهایی استخراج شده برای تصویر ورودی نشان داده شده است.

$$\boldsymbol{F_x} = < f_0, f_1, \dots, f_{P+1} > \qquad (24-2)$$

بردار ویژگی فوق الذکر معرف عددی بسیار خوبی از بافت تصویر مورد نظر است. در مرحله یافتن شبیه‌ترین تصاویر، می بایست فاصله بردار ویژگی تصویر پرس وجو با تصویر درون پایگاه سنجیده شده و در نهایت تصاویری از درون پایگاه که بیشترین شباهت (کم ترین فاصله) را با تصویر پرس و جو دارند انتخاب شوند.

## 2-6-5- الگوهای دودویی محلی بهبود یافته برای کاهش نویز

ما بر روی سازگار کردن خروجی الگوهای دودویی محلی با مسئله کاهش نویز تصاویر بافتی تحقیق کرده‌ایم. همانطور که پیش تر توضیح داده شد، در تصاویر بافتی، یک الگو (زیر بافت) خاص در تمام طول تصویر در حال تکرار شدن است . ما اعتقاد داریم که در تصاویر بافتی چنانچه قسمت هایی از تصویر دچار نویز گردد، با احتمال زیاد، بخش‌های مشابهی در درون همان تصویر می توان پیدا کرد که به بخش های نویزی شبیه بوده و دچار نویز نشده است . در راستای بررسی این موضوع، پیشنهاد می گردد که تصویر به پنجره هایی با ابعاد مساوی و بدون همپوشانی تقسیم گردد. سپس شبیه‌ترین پنجره‌ها به پنجره نویزی پیدا شود . در همین راستا مجدداً عملگر الگوهای دودویی محلی بر روی تمام پنجره ها اعمال شده و برای هر پنجره یک بردار ویژگی عددی طبق معادله (2-24) استخراج می گردد. در مرحله بعد می بایست فاصله بین بردارهای ویژگی استخراجی سنجیده شود . در این تحقیق انواع معیارهای فاصله و

معیارهای شباهت /عدم شباهت مورد آزمایش قرار گرفت . نتایج تجربی نشان داد که نسبت درستنمایی لگاریتمی [1] بالاترین کارایی [2] را فراهم آورد. نسبت درستنمایی لگاریتمی معیار عدم شباهت است که همواره مثبت و حداقل مقدار آن برابر صفر خواهد بود و کمینه‌شدن آن نشان-دهنده میزان شباهت با یک کلاس خاص می باشد. بنابراین کمترین مقدار محاسبه شده برای پنجره‌ها، به عنوان شبیه ترین پنجره و بزرگترین مقدار آن به عنوان حد آستانه شباهت بافت ها در بین پنجره ها در نظر گرفته می شود. نحوه محاسبه نسبت درستنمایی لگاریتمی دو بردار ویژگی در معادله زیر نشان داده شده است.

$$\boldsymbol{L} = (F_1, F_2) = \sum_{i=1}^{P+2} F_{1i} \log\left(\frac{F_{1i}}{F_{2i}}\right) \qquad (25-2)$$

در معادلات فوق، $F_1$ و $F_2$ به ترتیب بردارهای ویژگی استخراجی برای پنجره های 1 و 2 از تصویر است. همچنین "i" نشان دهنده بعد "i" ام از بردار ویژگی بوده و P نشان دهنده تعداد همسایگی در عملگر الگوی دودیی محلی استفاده شده می‌باشد.

[1] Log Likelihood Ratio
[2] Performance